\documentclass[10pt,twocolumn,letterpaper]{article}

\usepackage[pagenumbers]{cvpr} 

\usepackage[accsupp]{axessibility}  

\usepackage[accsupp]{axessibility}  
\usepackage{graphicx}
\usepackage{amsmath}
\usepackage{amssymb}
\usepackage{booktabs}
\usepackage[inline]{enumitem}
\usepackage{amsmath}
\usepackage{multirow}
\usepackage{multicol}
\usepackage[pagebackref,breaklinks,colorlinks]{hyperref}

\newcommand{\update}[1]{{#1}}

\usepackage[capitalize]{cleveref}
\crefname{section}{Sec.}{Secs.}
\Crefname{section}{Section}{Sections}
\Crefname{table}{Table}{Tables}
\crefname{table}{Tab.}{Tabs.}

\def\cvprPaperID{2229} 
\def\confName{CVPR}
\def\confYear{2022}

\newcommand{\twonorm}[2]{\left\lVert#1 - #2\right\rVert^{2}}
\newcommand{\twonormsum}[2]{\left\lVert#1 + #2\right\rVert^{2}}
\newcommand{\norm}[1]{\left\lVert#1\right\rVert^{2}}

\begin{document}

\title{Style-ERD: Responsive and Coherent Online Motion Style Transfer}

\author{Tianxin Tao$^{1}$ \hspace{10pt} Xiaohang Zhan \hspace{10pt}  Zhongquan Chen$^{2}$ \hspace{10pt}  Michiel van de Panne$^{1}$\\
$^{1}$ University of British Columbia \hspace{10pt} $^{2}$ University of California, Davis\\
{\tt\small \{taotianx,van\}@cs.ubc.ca \hspace{5pt} xiaohangzhan@outlook.com \hspace{5pt} czqchen@ucdavis.edu}
}
\maketitle

\begin{abstract}
   Motion style transfer is a common method for enriching character animation. Motion style transfer algorithms are often designed for offline settings where motions are processed in segments. However, for online animation applications, such as real-time avatar animation from motion capture, motions need to be processed as a stream with minimal latency. In this work, we realize a flexible, high-quality motion style transfer method for this setting. We propose a novel style transfer model, Style-ERD, to stylize motions in an online manner with an Encoder-Recurrent-Decoder structure, along with a novel discriminator that combines feature attention and temporal attention. Our method stylizes motions into multiple target styles with a unified model. Although our method targets online settings, it outperforms previous offline methods in motion realism and style expressiveness and provides significant gains in runtime efficiency.
\end{abstract}

\section{Introduction}
\label{sec:intro}

Animators commonly seek to create stylized motions to express the characters' personalities or emotions, thus making characters more lifelike. Since many computer animation techniques are based on motion capture data, the variety and diversity of the motion data play an essential role in the quality of the resulting animation. However, a capture-everything approach scales poorly if there is a need to capture every style, \eg, {\em childlike} or {\em depressed}, for every motion type. Hence, animators usually capture motion in a neutral style and then stylize them by hand, which is again laborious. This motivates automated methods for stylizing existing motions according to desired target-style labels.

\begin{figure}[t]
  \centering
   \begin{subfigure}{0.95\linewidth}
    \includegraphics[width=\linewidth]{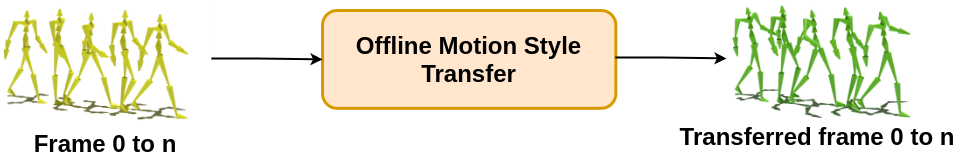}
    \caption{Offline motion style transfer.}
    \label{fig:short-a}
  \end{subfigure}
  \begin{subfigure}{0.95\linewidth}
    \includegraphics[width=\linewidth]{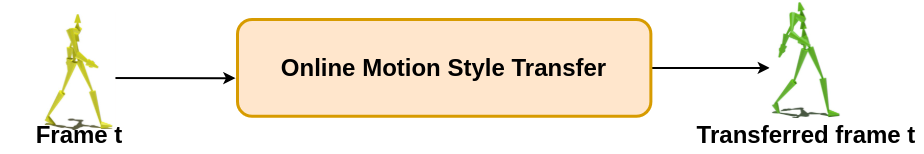}
    \caption{Online motion style transfer.}
    \label{fig:short-b}
  \end{subfigure}
   \caption{a) Offline motion style transfer processes motion segments, while b) online motion style transfer processes motions in a stream.}
   \label{fig:online_vs_offline}
\end{figure}

In this work, we develop a novel motion style transfer framework capable of stylizing streaming input motion data for online applications, which we define as \emph{Online Motion Style Transfer}.
As shown in Fig.~\ref{fig:online_vs_offline}, current motion style transfer methods~\cite{aberman2020unpaired, holden2016deep, holden2017fast, park2021diverse, dong2020adult2child, wen2021autoregressive} with deep learning models require a motion segment as input, and produce a transferred motion segment as output, and where the input segment has a minimum duration of 1\,s. 
While these methods make significant progress on a difficult problem, with a subset being described as being  real-time~\cite{xia2015realtime,smith2019efficient}, they still suffer from startup latency caused by waiting for the multiple frames required as input. For {\em online motion style transfer}, only the current frame is processed by the model, which enables the direct processing of the stream of motion data. We believe such transfer methods are more suitable for many novel applications requiring streaming motion data. For example, in animating a human avatar,  motion is captured online to animate the virtual avatar in real-time, and the streaming motion data needs to be processed with minimal latency. Online motion style transfer can also be easily incorporated into the workflows of real-time motion systems, such as games, interactive exhibitions, and augmented reality with minimal additional latency.

Motion style transfer exists as a long-standing research problem due to several difficulties, among many: 
\begin{enumerate*} [label=(\arabic*)] 
\item lack of a standardized qualitative style representation for motions,
\item difficulty in handling and generating temporally correlated data,
\item a lack of temporally registered motion data in different styles 
\end{enumerate*}. 
Several approaches~\cite{hsu2005style, shapiro2006style, yumer2016spectral, xia2015realtime} aim to solve this problem with manually designed models. However, they often fail to generalize well to large motion datasets with various styles. Researchers have developed more scalable methods with the rapid progress of deep learning machinery~\cite{holden2017fast, mason2018few, du2019stylistic, aberman2020unpaired}. However, only a few of them can transfer the motion to multiple target styles with a unified model~\cite{aberman2020unpaired, park2021diverse}. On top of the aforementioned challenges, {\em online} motion style transfer poses more difficulties because style and content are ill-defined and unrecognizable within one frame, yielding low-quality transfer results. Current offline motion style transfer methods are commonly conditioned on multiple input frames in order to understand the motion semantics, thus realizing better transfer at the cost of introducing non-trivial latency. \update{Although offline methods can be adapted to online settings by padding the current frame with past frames, there is no mechanism to guarantee the continuity among output frames.}



To accomplish high-quality, efficient motion style transfer with minimal latency, we embed knowledge regarding the previous frames in the memory of the motion transfer module in order to infer and track the style and content. The transfer module is thus aware of the context of content and style even when only presented with the current frame. 
We adapt the Encoder-Recurrent-Decoder~(ERD) framework to the online motion style transfer task by designing novel recurrent residual connections to capture features for each style. We name this novel architecture as \emph{Style-ERD}. In \emph{Style-ERD}, we enable each residual connection to learn its own initial hidden state $h_{0}$ conditioned on the style and content label. The learned hidden states are vital to the responsiveness of the style transfer results. In addition, to produce temporally coherent motions, we design a new discriminator with feature and temporal attention, \emph{FT-Att Discriminator}, to supervise the post-transfer style. As a result, our deep learning model demonstrates a strong capability to perform the desired motion style transfer efficiently and with minimal latency.

The contributions of this work are as follows:
\begin{enumerate*} [label=(\arabic*)] 
  \item We introduce the online motion style transfer problem and aim to stimulate future research into this area to facilitate real-time animation applications. \item We present a novel framework, \emph{Style-ERD}, as well as a new supervision module, \emph{FT-Att Discriminator}, achieving the goal of style transferring motion with minimum latency. Our style transfer framework provides a $5\times$ reduction in compute time, as compared with the current state-of-the-art approach. \item Our method can transfer the motion into its stylistic counterpart with high fidelity, showing better style transfer as compared to offline methods.
\end{enumerate*}
\section{Related Work}
\label{sec:rel}

\begin{figure*}[t]
  \centering
  \begin{subfigure}{0.6\linewidth}
    \includegraphics[width=\linewidth]{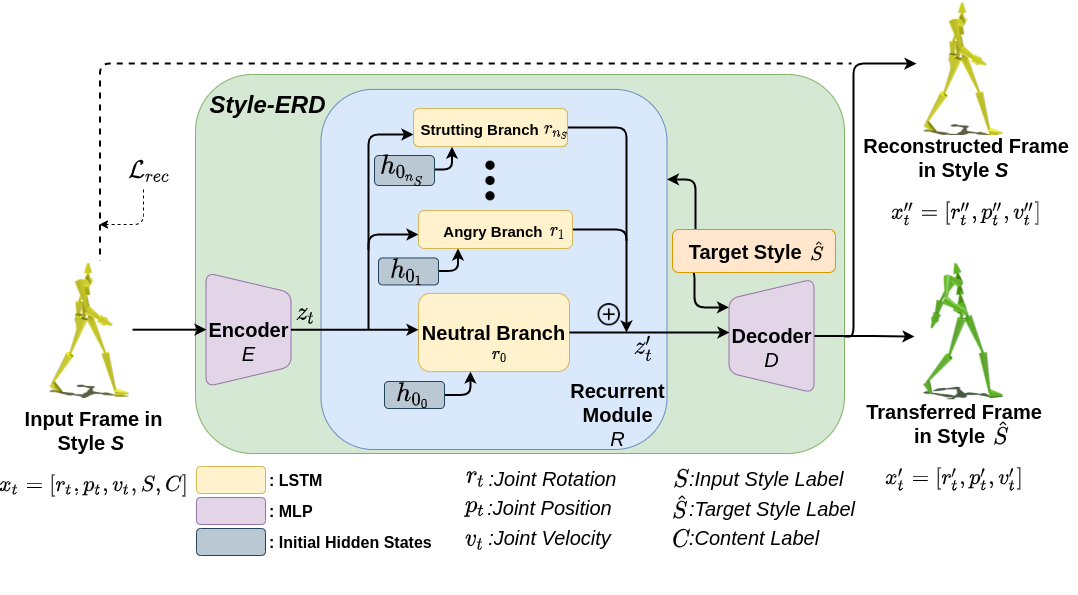}
    \caption{Style transfer module: \emph{Style-ERD}.}
    \label{fig:short-a}
  \end{subfigure}
  \hfill
  \vline
  \hfill
  \begin{subfigure}{0.35\linewidth}
    \includegraphics[width=\linewidth]{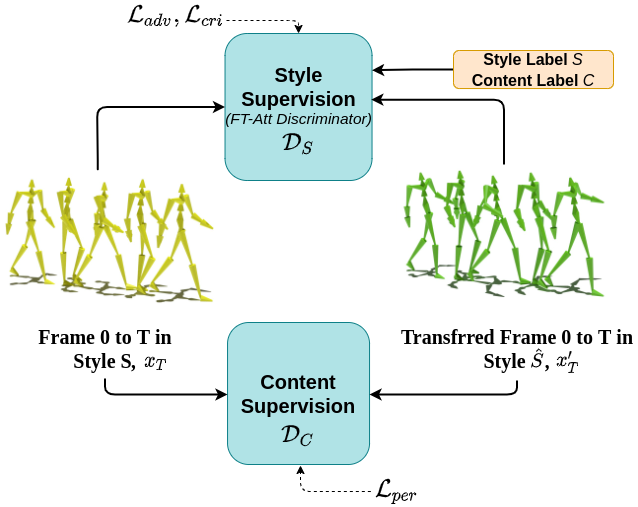}
    \caption{Style and content supervision modules.}
    \label{fig:short-b}
  \end{subfigure}
  \caption{Framework overview. (a) The input frame is encoded, forwarded in the recurrent module, and finally decoded to the original motion format in joint rotations. (b) At training time, the output motion segment is required to achieve the target style while preserving the content.}
  \label{fig:overview}
\end{figure*}

\subsection{Motion Synthesis and Control}
Motion synthesis has been a long-established research problem in computer animation and computer vision. Early methods are designed around motion graphs and search algorithms~\cite{motiongraphs2002, lee2002interactive, arikan2002interactive, kovar2004automated, lee2006precomputing, safonova2007construction, lo2008real, treuille2007near}, or Principle Component Analysis (PCA)~\cite{chai2005performance, tautges2011motion}. These methods are typically non-parametric in nature, in which case they demand large-and-complete datasets with limited  ability to generalize .


Recently, deep learning methods have also been applied to motion synthesis, motivated by their potential for scalability, generalization, and compute  efficiency. Holden \etal~\cite{holden2017phase} introduced a feedforward neural network model named Phase-Functioned Neural Networks~(PFNN) with a special weight blending mechanism. Further work built upon the weight blending mechanism of PFNN and improved its generalizability~\cite{zhang2018mode}, interaction~\cite{starke2019neural} and responsiveness~\cite{starke2020local}. 
Recurrent networks, including those enabled by Long Short-Term Memory~(LSTM) and variations thereof,  are another widely adopted structure for motion generation.
Fragkiadak \etal~\cite{fragkiadaki2015recurrent} used the ERD framework to predict the next pose given the current pose. Martinez \etal~\cite{martinez2017human} proposed a residual sequence-to-sequence architecture to learn differences between successive poses. The ERD framework was further extended to generate animation conditioned on keyframes~\cite{harvey2018recurrent,harvey2020robust}. Given a physics-based simulator, optimal control techniques~\cite{todorov2005generalized, pinneri2020sample, kim2021flexible} and reinforcement learning approaches~\cite{peng2018deepmimic, park2019learning, yin2021discovering, xie2020allsteps} also tackle the problem of motion generation with reference trajectories. 

\subsection{Image Style Transfer}

Style transfer for images was explored in~\cite{gatys2016image} through features extracted by convolutional neural networks. Johnson \etal~\cite{johnson2016perceptual} proposed a specific form of perceptual loss to accelerate the process. Later, instance normalization (IN) was proposed to normalize the style of image~\cite{ulyanov2016instance}, allowing the style to be manipulated by varying the mean and variance of IN~\cite{huang2017arbitrary}. Recently, impressive progress has been made towards enhancing picture quality~\cite{sanakoyeu2018style, zhang2019multimodal}, enabling user control~\cite{gatys2017controlling, risser2017stable}, improving runtime efficiency~\cite{li2016precomputed,li2018learning,gao2020fast} and accomplishing arbitrary style transfer~\cite{huang2017arbitrary,gu2018arbitrary,wu2021styleformer}.
In our work, the content supervision module is inspired by perceptual loss~\cite{johnson2016perceptual} and the style normalization effects introduced via IN~\cite{ulyanov2016instance}.



\subsection{Motion Style Transfer}
In early work, Hsu \etal~\cite{hsu2005style} proposed the use of a linear time-invariant~(LTI) model to represent motion style variance. \update{Wu \etal~\cite{wu2006line} further improved the performance of the LTI model by shortening the training time and simplifying the inputs.} Shapiro \etal~\cite{shapiro2006style} deployed Independent Component Analysis (ICA) to separate motions into different style components which can be adjusted to form stylized motions. Spectral domain features have also been used to capture the style differences between motions in a way that is largely invariant to the content~\cite{yumer2016spectral}. Xia \etal~\cite{xia2015realtime} proposed a local mixture of autoregressive models to extract the complex relationships between motion styles.

However, these models suffer from scalability issues and can still be slow for runtime use, motivating a recent focus on neural network methods. Holden \etal~\cite{holden2016deep,holden2017fast} propose a style transfer framework composed of a pretrained motion manifold to supervise content and Gram matrices to represent the style of the motion. 
In~\cite{smith2019efficient}, the authors build a pose network, a foot contact network and a timing network with feedforward layers to achieve outstanding runtime efficiency for motion style transfer. 
Aberman \etal~\cite{aberman2020unpaired} adopt the IN together with the AdaIN layer applied in image style transfer to the motion style transfer task, \update{and propose a transfer algorithm without requiring paired motion data.} Inspired by that, Park \etal\cite{park2021diverse} replace the 1D temporal convolution structure with a spatial-temporal graph
convolution. A residual model can also be used to extract the style ingredients of the motion~\cite{mason2018few}. Wen \etal~\cite{wen2021autoregressive} propose a style transfer framework with generative flow.

Inspired by ~\cite{holden2017fast, aberman2020unpaired}, our framework adopts the two-way constraints on style and content. In contrast to those previous deep learning models, our framework can operate with a single input frame while delivering high-quality motion in a target style and being considerably more computationally efficient in an online setting.
\section{Methodology}
\label{sec:methodology}

Our goal is to develop an online motion style transfer algorithm with high-quality transfer and minimal latency. In particular,
we seek to reduce the number of input frames needed at each timestep to synthesize the current frame of the target style.
However, with fewer input frames, the style transfer model may err in interpreting the style and content. 
We therefore leverage a recurrent model to maintain relevant estimates of the style and content. 
Our framework consists of three components: a style transfer module, a style supervision module and a content supervision module. 
An overview of our method is displayed in Fig.~\ref{fig:overview}.

Inspired by the ERD framework~\cite{fragkiadaki2015recurrent}, we name the style transfer module as \emph{Style-ERD}.
It is characterized by multiple recurrent residual connections and by hidden states that have learned initial values which are conditioned on the input. The novel recurrent residual connections play a key role in the success of our method because the memory of past frames provides style and content information regarding the current frame while the residuals capture the features of each style. The \emph{Style-ERD} model achieves the goal of style transfer from each single frame input in real-time.

The style transfer module (\emph{Style-ERD}) delivers poor style transfer when used with only a reconstruction loss. Conditioning on style and content before and after the transfer task can boost the style transfer effects. Both supervision modules take multiple frames of motion as input. This multi-frame input to supervision modules does not hinder the online property of our method since the supervision modules are unused at inference time. We propose a novel attention mechanism that spans both feature space and temporal space of the feature maps in the style discriminator, \emph{FT-Att Discriminator}, to enable the style transfer module to avoid mode-collapse issues that would otherwise preclude modeling the desired variety of style and content. The content supervision module adopts the idea of perceptual loss~\cite{johnson2016perceptual} with features that focus on the content of motion. 

\subsection{Architecture}

\begin{figure}[t]
  \centering
  \includegraphics[width=0.92\linewidth]{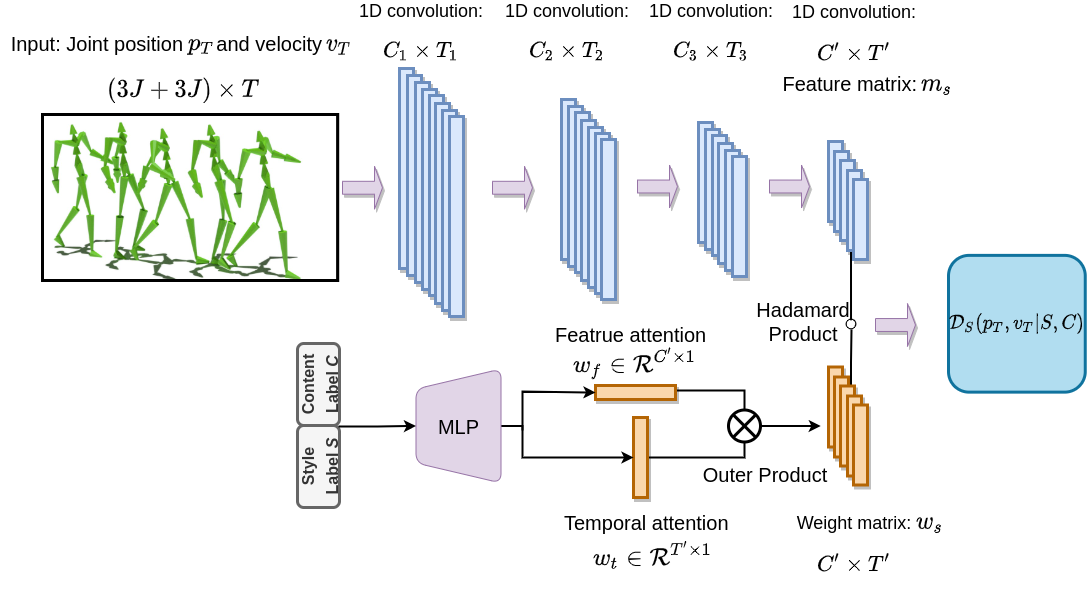}

  \caption{\emph{FT-Att Discriminator} structure. The discriminator forms the weight matrix via an outer product between two attention vectors, then applies the attention matrix on the extracted features via a Hadamard product.}
  \label{fig:discriminator}
\end{figure}

\noindent\textbf{Motion Transfer Module.} Our style transfer module, \emph{Style-ERD}, consists of three parts: an encoder $E$ to compress the input frame $x_{t}$, a recurrent module $R$ consisting of residual connections to learn the offsets of different styles, and a decoder $D$ to map the latent code back to the transferred motion frame $x'_{t}$ represented by joint rotations.

The input frame $x_{t}$ contains joint rotations in unit quaternions $r_{t} \in  \mathbb{R}^{4 \times J} $, joint positions offset by the root $p_{t} \in \mathbb{R}^{3 \times J}$, and linear joint velocities $v_{t} \in \mathbb{R}^{3 \times J}$ at timestep $t$, where $J$ is the number of joints. Additionally, the encoder is conditioned on the style label $S$ and content label $C$ of the input frame $x_{t}$ while the decoder is conditioned on the target style label, $\hat{S}$. Both style and content labels are represented by one-hot vectors.

The encoder consists of a two-layer MLP (multilayer perceptron) to compress the input to a low-dimensional space, $z$. We choose to compress the input for two reasons: \begin{enumerate*} [label=(\arabic*)] \item A low-dimensional latent space can simplify capturing an abstracted style representation; and \item With this low-dimensional bottleneck and the given training tasks, the encoder can normalize the style of the input frame to neutral.\end{enumerate*}

The recurrent module is designed as a stack of LSTM layers, \ie,~$ R = [r_{0}, r_{1}, \dots, r_{n_{S}}]$, with each one learning a style offset of one specific style with respect to the neutral style. Here, we assign the recurrent branch $r_{0}$ to learn the features of the neutral style, which serves as the basis for all other style offsets. Then, the residual value computed by the target style branch $r_{\hat{S}}(z_{t})$ is added to the neutral branch output, $r_{0}(z_{t})$. Thus, the operation of our recurrent module can be expressed as: $z'_{t} = r_{0}(z_{t}) + r_{\hat{S}}(z_{t})$

In addition, it is challenging to perform style transfer on the first few frames because the memory of the LSTM layers may not have seen enough information to infer the necessary style information. Common ways to initialize hidden states include setting the hidden states to zeros~\cite{li2017auto} or random noise~\cite{zimmermann2012forecasting}, and treating the initial hidden states as parameters for the network to learn~\cite{harvey2018recurrent}. In order to enhance the performance on the first few frames, we propose learning multiple initial states $h_{0} = [h_{0_{0}}, h_{0_{1}},\dots,h_{0_{n_{S}}}]$ conditioned on the style label $S$ and the content label $C$. Specifically, assume there are $n_{C}$ different content labels; the neutral branch $r_{0}$ learns $n_{C}$ initial hidden states simultaneously and selects the one corresponding to the content label. Similarly, each style branch learns its own initial hidden state for its corresponding style.

The conditional decoder $D$ expands the latent code $z'_{t}$ back to joint rotations in quaternions and linear joint velocities through four MLP layers, further conditioned on the target style label $\hat{S}$. Joint positions are also computed via forward kinematics of joint rotations. 

\noindent\textbf{Style Supervision Module.} We propose a novel discriminator $\mathcal{D_{S}}$ with an attention mechanism, \emph{FT-Att Discriminator}, to supervise the style transfer task. Fig.~\ref{fig:discriminator} shows the structure of \emph{FT-Att Discriminator}. Unlike the style transfer module, our discriminator receives a segment of $T$ ($T = 24$) frames as input to infer the style of the input motion. Each frame is represented by the concatenation of joint positions offset by the root $p_{t} \in \mathbb{R}^{3 \times J}$ and linear joint velocities $v_{t} \in \mathbb{R}^{3 \times J}$ as Aberman \etal~\cite{aberman2020unpaired} found positions are more representative for styles than rotations.

The discriminator attempts to distinguish generated motions from real motion samples according to style labels $S$ and content labels $C$. We adopt a 1D temporal convolution structure similar to~\cite{liu2019few,aberman2020unpaired} to extract the 2D feature matrix $m_s \in \mathbb{R}^{C' \times T'}$ but add novel attention modules conditioned on style and content. The intuition behind the attention modules is that the discriminator should judge motion style according to the desired style and its content by weighing the features unevenly. The attention modules are comprised of MLP layers with style and content labels as input, and outputs feature attention vector $w_{f} \in \mathbb{R}^{C'}$ and temporal attention vector $w_{t} \in \mathbb{R}^{T'}$. We then compute an outer product between the feature attention $w_{f}$ and temporal attention $w_{t}$ to form a weight matrix $w_{s}\in \mathbb{R}^{C' \times T'}$: $w_{s} = w_{f} \bigotimes w_{t}$. Finally, the weight matrix $w_{s}$ is applied to the feature matrix $m_{s}$ via a Hadamard product, \ie,~element-wise multiplication. Thus, given the feature map $m_{s}$ and weight matrix $w_{s}$, the output of the discriminator can be expressed as:
\begin{equation}
  \mathcal{D}_{S}(p_{T}, v_{T}|S,C) = \sum_{i=0}^{C'}\sum_{j=0}^{T'}(m_{s} \circ w_{s})[i, j]
  \label{eq:dis_op}
\end{equation}

\noindent\textbf{Content Supervision Module.} At the same time as transferring style, we expect the content of the motion to be unaltered. We apply perceptual loss~\cite{johnson2016perceptual} based on a pre-trained content-classification network $\mathcal{D}_{C}$ to preserve content. The classification network follows the same convolution layers as the discriminator while accepting joint rotations, joint positions and velocities as input. Inspired by the style normalization effects of IN~\cite{huang2017arbitrary, aberman2020unpaired}, each convolution layer is followed by IN such that the classification network focuses on the motion content and disregards the style.

\subsection{Training}
The training process is analogous to the training of the standard Generative Adversarial Network~\cite{goodfellow2014generative}. As a generator, the proposed style transfer module \emph{Style-ERD} is trained to reconstruct the input frame and to stylize the frame to the target style to fool the discriminator, while the objective of the \emph{FT-Att Discriminator} is to distinguish the transferred motions from real data samples. We add perceptual loss and further adopt a gradient penalty in the discriminator to improve the overall training process. For simplicity and clarity, we use the notation (.)' to indicate attributes of the transferred results.

\noindent\textbf{Reconstruction.} Given a motion input $x_{t}$ and a target style label $\hat{S}$ identical to the original style, the motion transfer module should output an identical frame $x''_{t} = [r''_{t}, p''_{t}, v''_{t}]$. This reconstruction task can be viewed as an auxiliary task to learn disentangled style variance for each residual branch. The reconstruction loss is applied over the joint rotations in quaternions $r_{t}$, translational joint positions $p_{t}$ and velocities $v_{t}$:
\begin{equation}
    \mathcal{L}_{quat}(r_{t}, r''_{t}) = \norm{cos^{-1}(\lvert r_{t} \cdot r''_{t} \rvert)},
    \label{eq: quat}
\end{equation}
\begin{equation}
    \mathcal{L}_{rec_{t}} = \mathcal{L}_{quat}(r_{t}, r''_{t}) + \frac{1}{2}\twonorm{p_{t}}{p''_{t}} + \twonorm{v_{t}}{v''_{t}},
    \label{eq: rec}
\end{equation}
where $\mathcal{L}_{quat}$ denotes the quaternion difference represented by the angle between two rotations in radians. More details about $\mathcal{L}_{quat}$ can be found in the supplementary material.\\

\noindent\textbf{Style Transfer.} We adopt the Least Squares Generative Adversarial Networks (LSGAN) \cite{mao2017least} framework to train the \emph{FT-Att Discriminator}. We assume that neutral style motion serves as a common basis for other styles. At training time, we set the target styles of all neutral motions to be any other existing style in the dataset, while motions in other styles except neutral should be transferred to the neutral style. With these training objectives, we expect the encoder $E$ to normalize the input motion to a neutral style. Therefore, the adversarial loss is applied to manipulate the style of the motion by fooling the critic. At the same time, the critic is trained to distinguish the fake generated motion from the real motion sample:
\begin{equation}
    \mathcal{L}_{adv} = \norm{\mathcal{D}_{s}(p'_{T},v'_{T}|\hat{S},C)},
    \label{eq: adv}
\end{equation}
\begin{equation}
    \begin{split}
        \mathcal{L}_{cri}   = & \twonorm{\mathcal{D}_{s}(p_{T},v_{T}|S,C)}{1}\\
                              & + \twonormsum{\mathcal{D}_{s}(p_{T}',v_{T}'|\hat{S},C))}{1}.
    \end{split}
    \label{eq: cri}
\end{equation}

\noindent\textbf{Gradient Penalty.} GAN training is known to suffer from instability and convergence issues, with multiple approaches proposed to address this issue~\cite{salimans2016improved,hazan2017adversarial,mescheder2018training,heusel2017gans,karras2017progressive}. In this work, we apply a gradient penalty on the real samples to prevent the discriminator from creating a non-zero gradient orthogonal to the data manifold when the generator produces the true data distribution~\cite{mescheder2018training}:
\begin{equation}
    \mathcal{L}_{gp} = \norm{\nabla_{\hat{x}}\mathcal{D}_{s}(\hat{x})|_{\hat{x}=(p_{T},v_{T}|S,C)})}
    \label{eq: gp}
\end{equation}

\noindent\textbf{Perceptual Loss.} In order to preserve content before and after the transfer, we add a perceptual loss~\cite{johnson2016perceptual} $\mathcal{L}_{per}$ to the generator with a pretrained multi-class content-classification network $\mathcal{D}_{C}$. The perceptual loss encourages the convolution feature maps $\phi$ extracted by the classification network before and after the transfer to match:
\begin{equation}
    \mathcal{L}_{per} = \twonorm{\phi}{\phi'}.
    \label{eq: per}
\end{equation}

The final loss applied to the motion style transfer module (generator) is a weighted sum of reconstruction, adversarial, perceptual loss while a gradient penalty is added to the discriminator loss:
\begin{equation}
    \mathcal{L}_{gen} = \sum_{t=0}^{T}\mathcal{L}_{rec_t} + w_{adv}\mathcal{L}_{adv} + w_{per}\mathcal{L}_{per},
    \label{eq: gen}
\end{equation}
\begin{equation}
    \mathcal{L}_{dis} = \mathcal{L}_{cri} + w_{gp}\mathcal{L}_{gp},
    \label{eq: dis}
\end{equation}
where we set $w_{adv}= 1$, $w_{per} = 0.1$ and $w_{gp} = 128$.
\section{Experiments}
\label{sec:experiments}

\begin{figure*}
  \centering
  \footnotesize
  \begin{tabular}{l|ccc}
         & Neutral run into angry style & Neutral kick into sexy style & Neutral jump into childlike style\\
         
         \midrule
         
        Input Motion&
        \begin{subfigure}{0.25\textwidth}
            \includegraphics[width=\linewidth]{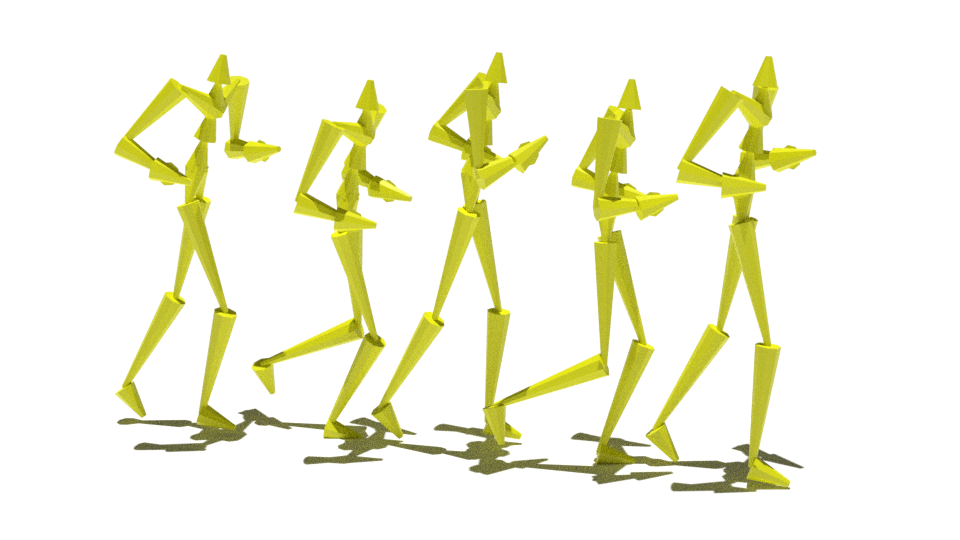}
        \end{subfigure}&
        \begin{subfigure}{0.25\textwidth}
            \includegraphics[width=\linewidth]{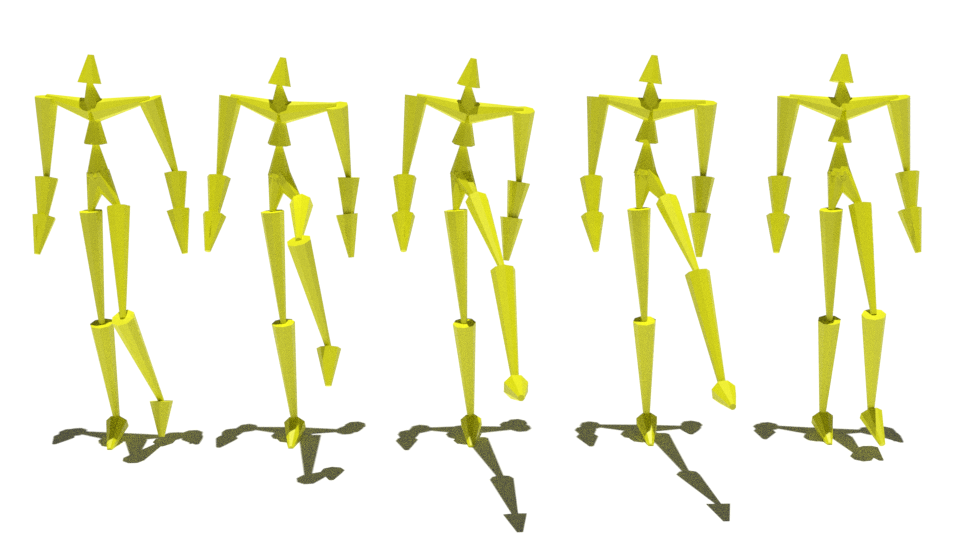}
        \end{subfigure}&
        \begin{subfigure}{0.25\textwidth}
            \includegraphics[width=\linewidth]{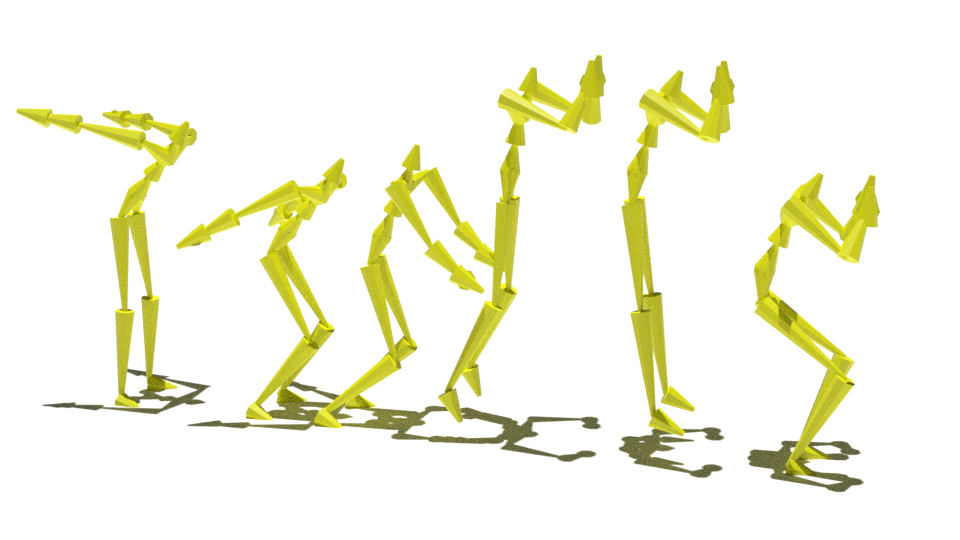}
        \end{subfigure}\\
        
        \midrule
        
        Ours&
        \begin{subfigure}{0.25\linewidth}
            \includegraphics[width=\linewidth]{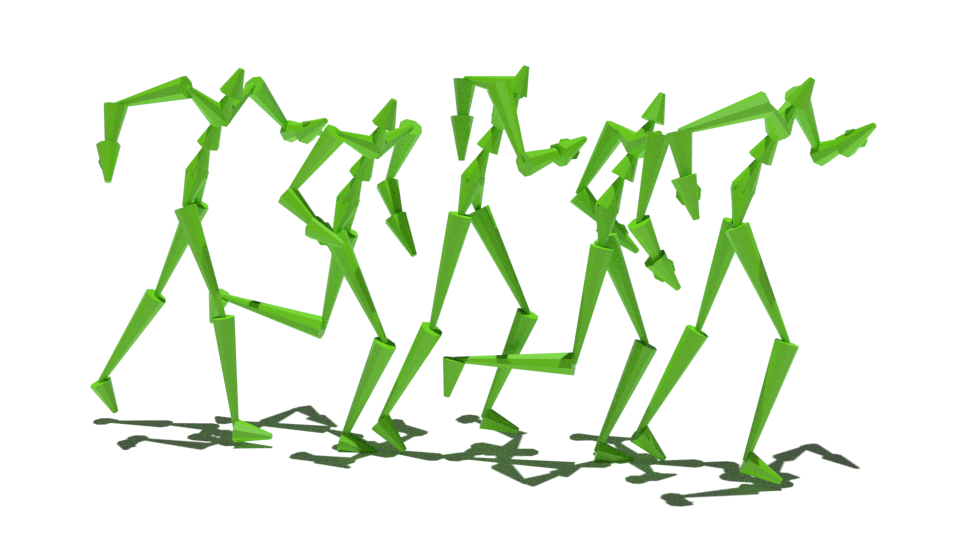}
        \end{subfigure}&
        \begin{subfigure}{0.25\linewidth}
            \includegraphics[width=\linewidth]{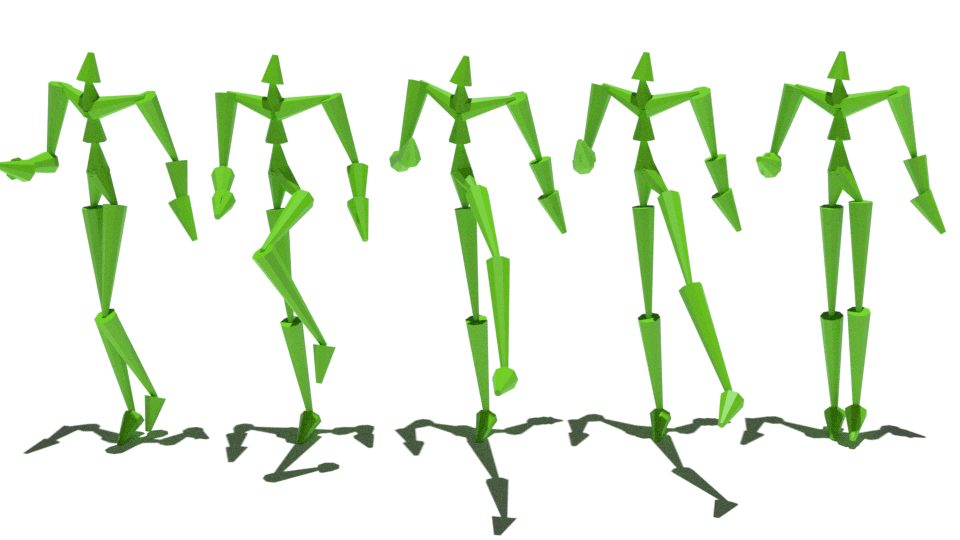}
        \end{subfigure}&
        \begin{subfigure}{0.25\linewidth}
            \includegraphics[width=\linewidth]{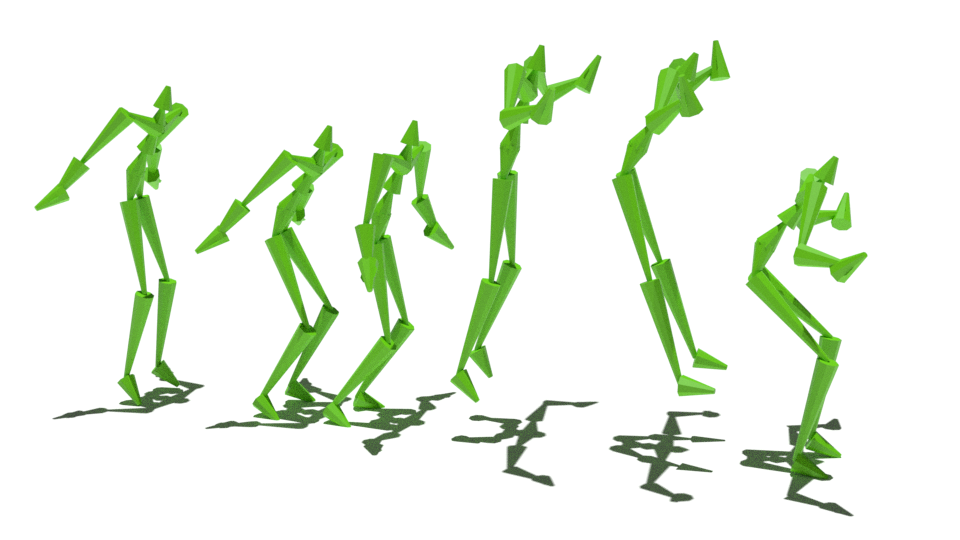}
        \end{subfigure}\\
        
         \midrule
        
        Aberman \etal~\cite{aberman2020unpaired}&
        \begin{subfigure}{0.25\linewidth}
            \includegraphics[width=\linewidth]{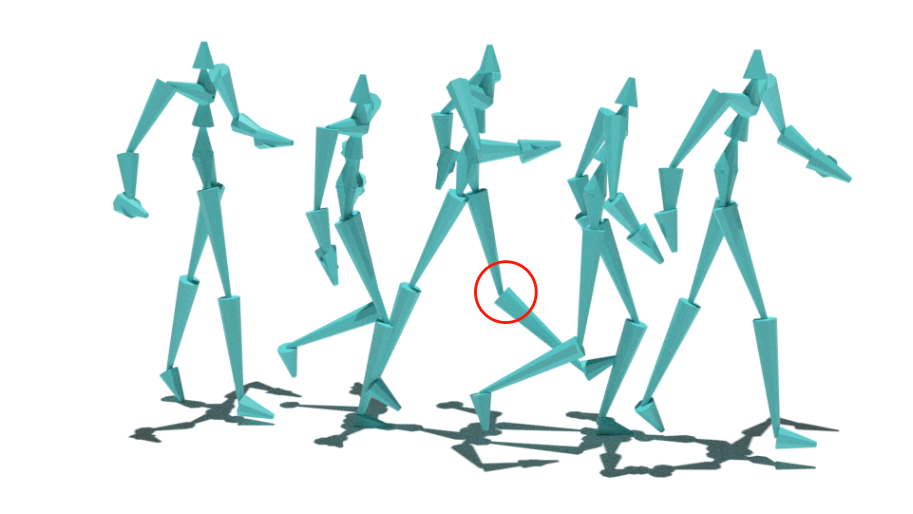}
        \end{subfigure}&
        \begin{subfigure}{0.25\linewidth}
            \includegraphics[width=\linewidth]{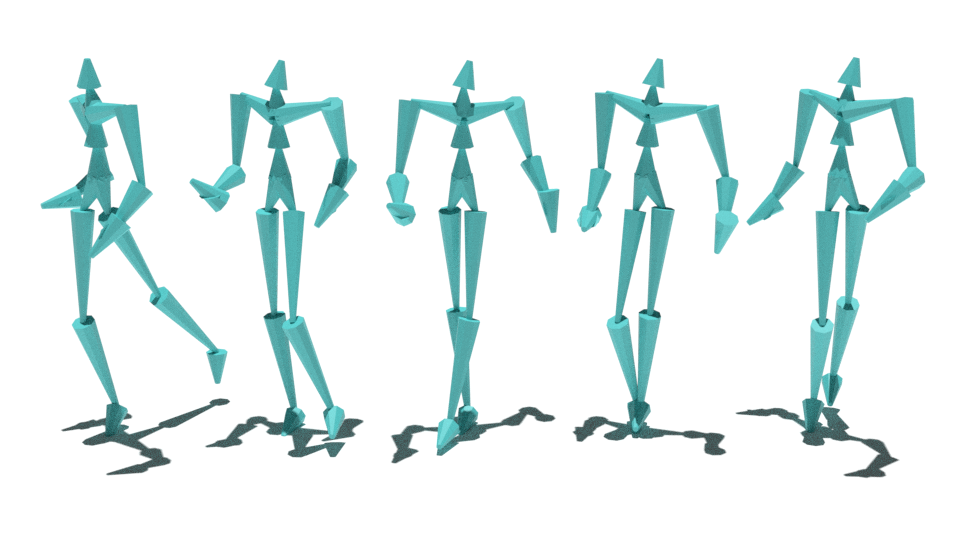}
        \end{subfigure}&
        \begin{subfigure}{0.25\linewidth}
            \includegraphics[width=\linewidth]{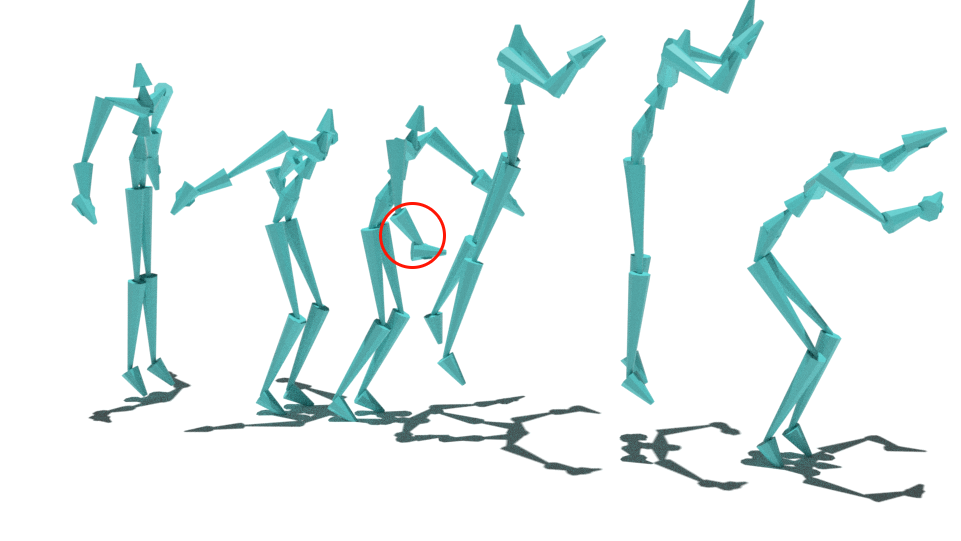}
        \end{subfigure}\\
        
         \midrule
        
        Park \etal~\cite{park2021diverse}&
        \begin{subfigure}{0.25\linewidth}
            \includegraphics[width=\linewidth]{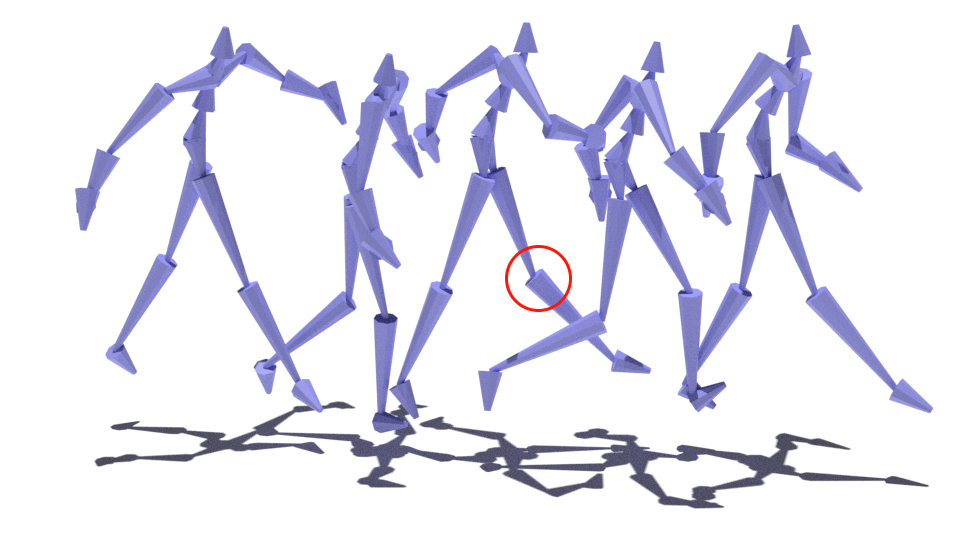}
        \end{subfigure}&
        \begin{subfigure}{0.25\linewidth}
            \includegraphics[width=\linewidth]{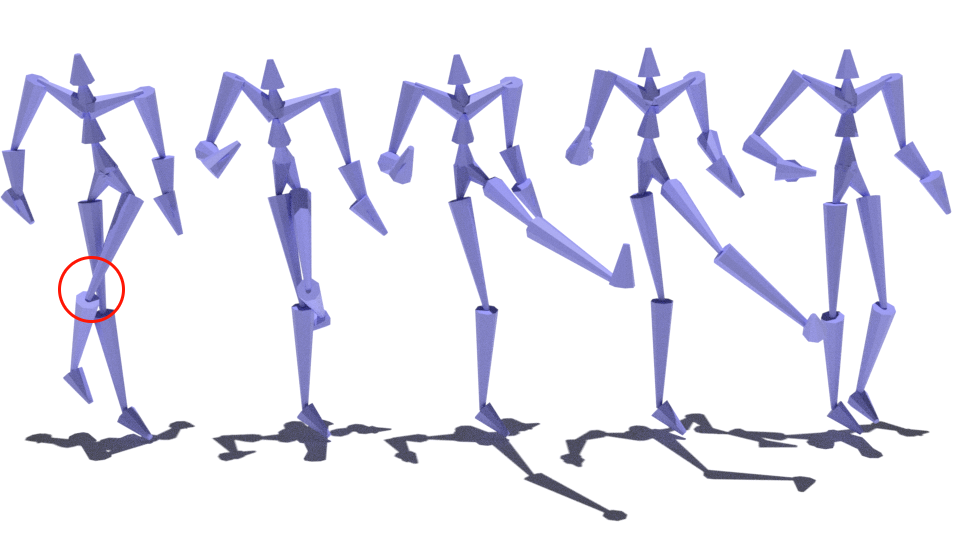}
        \end{subfigure}&
        \begin{subfigure}{0.25\linewidth}
            \includegraphics[width=\linewidth]{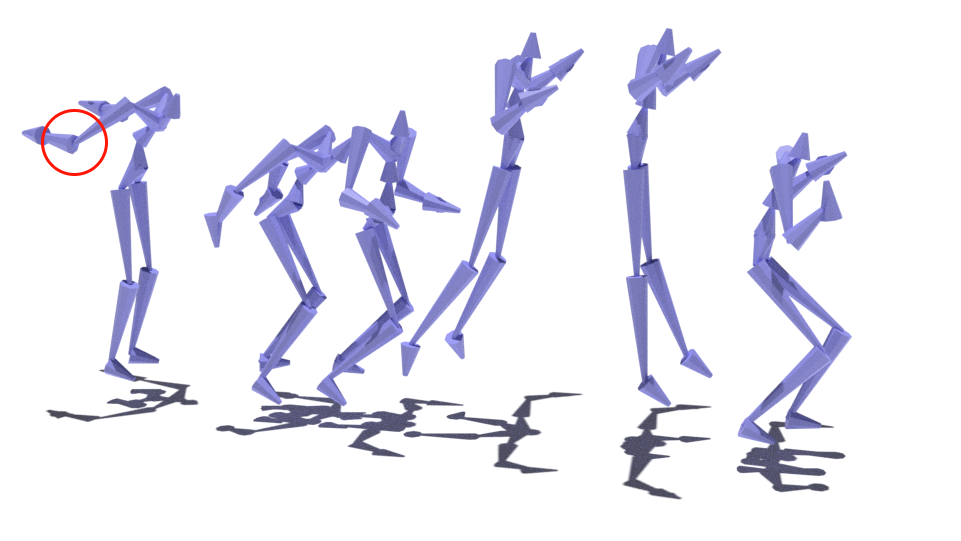}
        \end{subfigure}\\
        
        \midrule
        
        Style Reference&
        \begin{subfigure}{0.25\linewidth}
            \includegraphics[width=\linewidth]{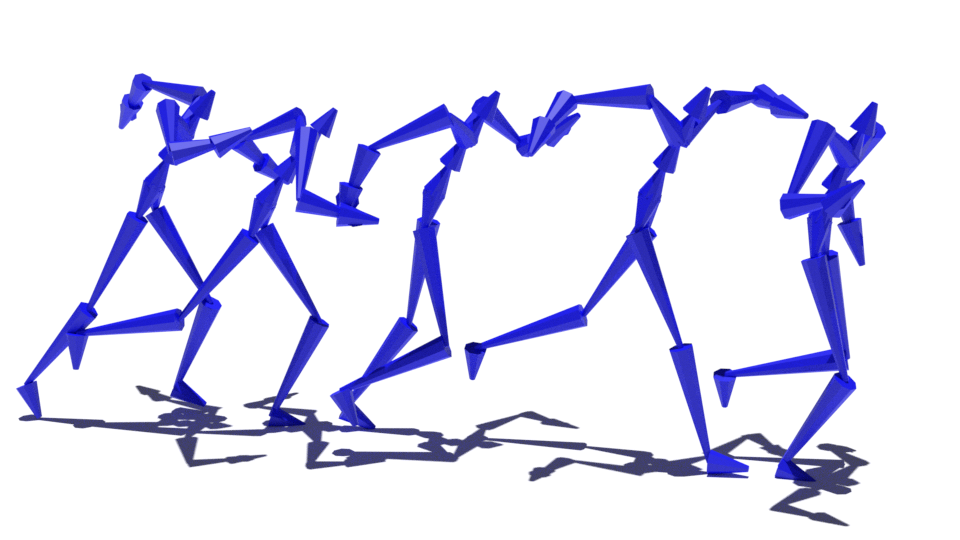}
            \caption{}
            \label{fig:comparison-a}
        \end{subfigure}&
        \begin{subfigure}{0.25\linewidth}
            \includegraphics[width=\linewidth]{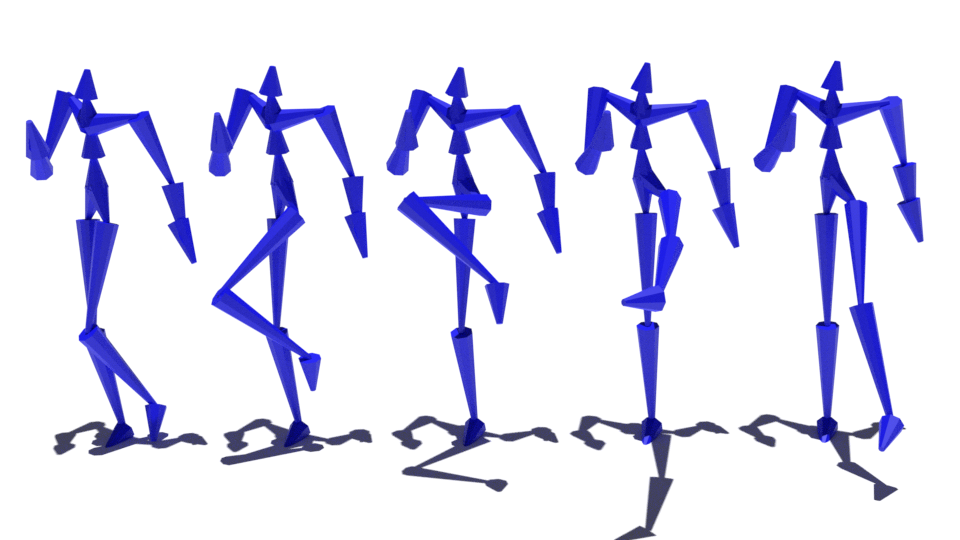}
            \caption{}
            \label{fig:comparison-b}
        \end{subfigure}&
        \begin{subfigure}{0.25\linewidth}
            \includegraphics[width=\linewidth]{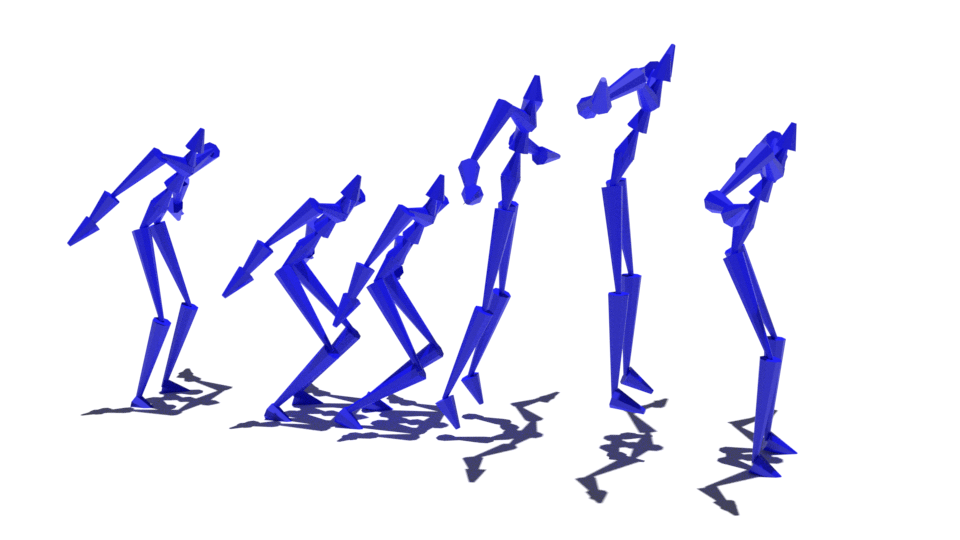}
            \caption{}
            \label{fig:comparison-c}
        \end{subfigure}
        
  \end{tabular}
 \caption{Style transfer comparison: show the input motion, style transfer results produced by our method, the methods in Aberman \etal~\cite{aberman2020unpaired} and in Park \etal~\cite{park2021diverse}. Example artifacts are circled in red. Style reference is the existing input motion content in the target style, as provided by the test set. Ideally, transferred results should resemble the style references, with the motion content remaining unchanged.}
 \label{fig:comparison}
\end{figure*}

We test our framework based on the dataset provided by Xia \etal~\cite{xia2015realtime}. We first compare the style transfer results with previous offline style transfer methods~\cite{aberman2020unpaired,park2021diverse}. We adopt the quantitative Fréchet Motion Distance (FMD) metric proposed in~\cite{park2021diverse} to evaluate the quality of the transferred results, which is a variant of the Fréchet Inception Distance (FID)~\cite{heusel2017gans}. We train a denoising autoencoder as the feature extractor for FMD. Additionally, we perform qualitative evaluations on the transferred results according to three criteria: style expressiveness, temporal consistency and content preservation. \update{Furthermore, we conduct a user study to evaluate the quality of the transfer results, which is described in the supplementary material.} Second, we measure the runtime efficiency of our method, to assess feasibility for use in real-time online applications. We also demonstrate that insufficient input frames can degrade the results of offline style transfer methods~\cite{aberman2020unpaired,park2021diverse}, and that our approach works well in the online transfer setting, with minimal latency. We then experiment with interpolation of the style ingredients of the transfer module. Finally, we test the generalization property of our methods. For better visualization and comparison, we refer readers to our supplementary video and material.

\subsection{Dataset}

Our method is trained and tested on the dataset supplied by Xia \etal~\cite{xia2015realtime}. The motion clips cover $n_{S}=7$ different styles and $n_{C}=5$ distinct types of content. We downsample the original 120fps motion data to 60fps, which results in around 1500 motion clips in total. Each motion clip spans between 28 to 80 frames. To ensure that the model is agnostic to the choice of initial frame, we further split each motion clip into multiple windows of $T=24$ frames, with an overlap of 4 frames. The motion clips are assigned randomly into the training and test sets without any overlap.

\subsection{Style Transfer Quality}


\begin{table}
  \scriptsize
  \centering
  \begin{tabular}{lc|lc}
    \toprule
    Method & FMD$\downarrow$ & Ablation study & FMD$\downarrow$\\
    \midrule
    Aberman \etal~\cite{aberman2020unpaired} & 563.41 & Ours w/ $\mathcal{L}_{gen}=\mathcal{L}_{rec}$ & 285.17 \\
    Park \etal~\cite{park2021diverse} & 190.94 & Ours w/ $\mathcal{L}_{gen}=\mathcal{L}_{rec}+\mathcal{L}_{adv}$ & 75.40\\
    Ours & \textbf{61.95} & Ours w/ $\mathcal{L}_{gen}=\mathcal{L}_{rec}+\mathcal{L}_{per}$ & 380.78\\
     & & Ours w/o attention & 382.57\\
     & & Ours w/o learnt initial states & 309.70\\
    \bottomrule
  \end{tabular}
  \caption{Quantitative evaluations on our method and existing motion style transfer methods, as well as the ablation study results.
  }
  \label{tab:fmd}
\end{table}

Tab.~\ref{tab:fmd} lists the FMD scores for our method and alternative approaches~\cite{aberman2020unpaired,park2021diverse}. Our method achieves the lowest FMD, which is indicative of its transferred results being closer to real stylized motion samples, in a probabilistic sense, than the alternative methods.  Fig.~\ref{fig:comparison} shows three sets of style transfer results using our method and those proposed by Aberman \etal~\cite{aberman2020unpaired} and Park \etal~\cite{park2021diverse}. In addition to the above objective measures, the motions transferred with our method subjectively resemble the existing motions in the target styles while Aberman \etal~\cite{aberman2020unpaired}'s method, which is an unsupervised model, may exhibit a limited change of style (see Fig.~\ref{fig:comparison-c}). Furthermore, our approach produces consistent style effects throughout the motion cycle, while results produced by \cite{park2021diverse} may contain artifacts at the beginning and the end of the motion (see Fig.~\ref{fig:comparison-b}). In terms of content preservation, the content of the transferred motion by our method is easily recognizable and remains identical to the input. In contrast, the motions produced by~\cite{aberman2020unpaired} and~\cite{park2021diverse} exhibit a degree of content variation after the style transfer (see Fig.~\ref{fig:comparison-a} and Fig.~\ref{fig:comparison-b}). Our framework can also stylize heterogeneous action sequences, \ie,~one motion clip with multiple content types. Results are included in the supplementary material.

\subsection{Efficiency}
\begin{figure}
  \centering
    \begin{subfigure}{0.3\linewidth}
        \includegraphics[width=\linewidth]{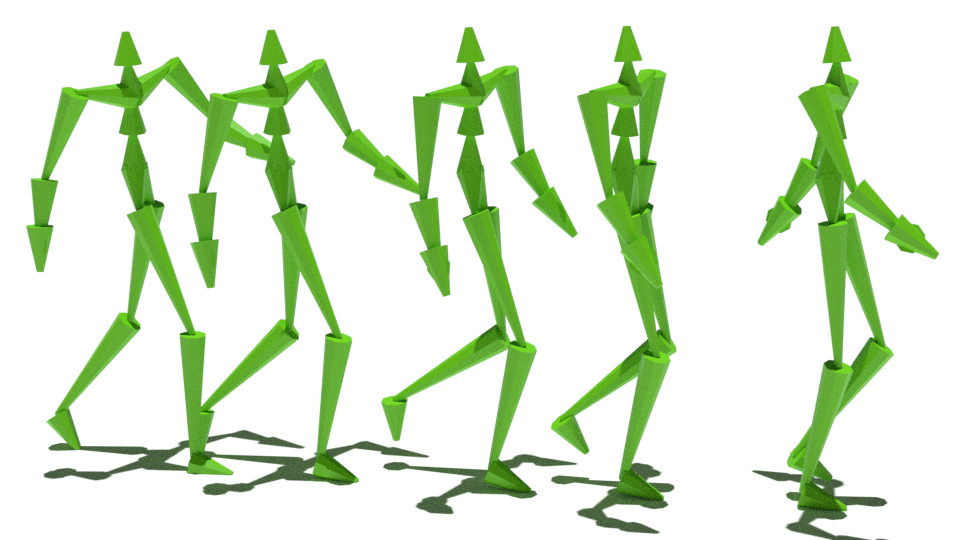}
        \caption{Our method}
    \end{subfigure}
    \begin{subfigure}{0.3\linewidth}
        \includegraphics[width=\linewidth]{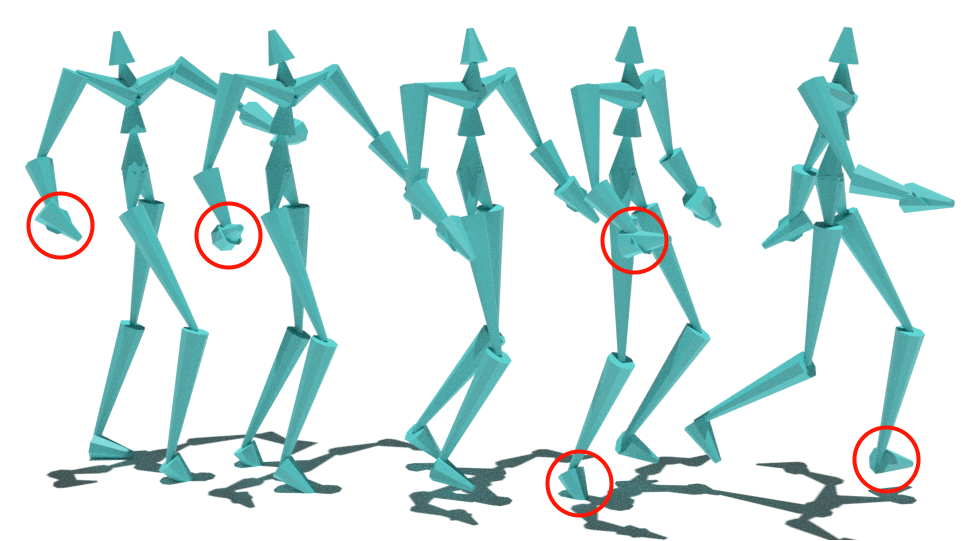}
        \caption{Aberman \etal~\cite{aberman2020unpaired}}
    \end{subfigure}
    \begin{subfigure}{0.3\linewidth}
        \includegraphics[width=\linewidth]{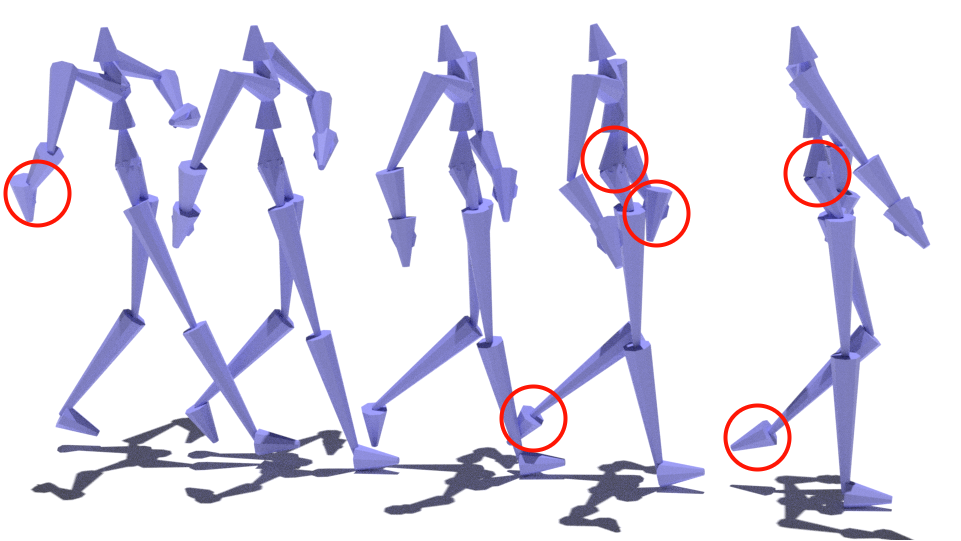}
        \caption{Park \etal~\cite{park2021diverse}}
    \end{subfigure}
    \caption{Style transfer results of the first ten frames. Task: Neutral walk into proud style. Noticeable artifacts of awkward joint rations are circled in red.}
    \label{fig:tenframe}
\end{figure}

We evaluate our method in an online setting based on two criteria: \begin{enumerate*} [label=(\arabic*)] \item startup latency and \item runtime\end{enumerate*}. Startup latency is important in an online motion transfer task because the delay can directly impact user interaction. In Tab.~\ref{tab:runtime}, we list the runtime of our method and alternative approaches~\cite{park2021diverse,aberman2020unpaired,xia2015realtime}, and the designed input length at training. Although those methods are not restricted to a particular length of input, we find they often have degraded motion transfer when used with fewer input frames. Thus, the designed input frames serve as an indicator for latency in a streaming application. In Fig.~\ref{fig:tenframe}, the motion transfer results by~\cite{aberman2020unpaired,park2021diverse} contain noticeable artifacts on the rotation of the limbs when only 10 frames are supplied as content input. On the other hand, our framework can produce good motion transfer results from the very beginning due to the learned initial hidden states and recurrent structure.

We also measure the runtime for style transfer, per frame, in an online setting and provide additional measurements on multiple frames as reference. The runtime performance is tested on a PC with NVIDIA GeForce GTX 1060 GPU (6GB), with the exception of the measurements for~\cite{xia2015realtime}, for which use their documented results. Since previous methods~\cite{aberman2020unpaired,park2021diverse} are not designed for online purposes, we pad the current frame with the past 32 frames in the online runtime tests. In terms of runtime, our method significantly outperforms the alternative transfer methods, providing a speedup of a factor of 5 as compared to  the second fastest runtime of Park \etal\cite{park2021diverse}, in the online setting. A shorter input window enables lower latency of handling streaming motion data to produce realistic transfer results. Despite the method introduced by Xia \etal~\cite{xia2015realtime} requiring the second shortest input window, their algorithm operates much slower than ours with the default $k$-nearest neighbors ($k$NN) implementation. Our method also achieves the fastest runtime when transferring short motion segments. However, the sequential nature slows down the runtime of our method when used in conjunction with a long input sequence, although this is not our focus. In conclusion, our method outperforms previous style transfer methods by a large margin in both runtime and startup latency. Our method is sufficiently fast for current online streaming applications at up to 120 Hz.

\begin{table}
  \scriptsize
  \centering
  \begin{tabular}{l|cccc|c}
    \toprule
    \multirow{2}{*}{Method} & \multicolumn{4}{c}{Runtime (ms)} & Input Length\\
                            & \textbf{Online} & L=60 & L=90 & L=120 & (frames) \\
    \midrule
    Aberman \etal~\cite{aberman2020unpaired} & 26.82 & 27.02 & 27.56 &27.74 & 32 \\
    Park \etal~\cite{park2021diverse} & 10.21 & 10.20 & 10.24 & \textbf{10.25} & 64\\
    Xia \etal~\cite{xia2015realtime} & 18.10 & 217.2 & 325.8 & 434.4 & 5\\
    Ours & \textbf{1.73} & \textbf{7.47} & \textbf{10.18} & 13.22 & \textbf{1} \\
    \bottomrule
  \end{tabular}
  \caption{Runtime measurements in an online setting and different input frame number L. Input length indicates the ideal length of input segment, which relates to latency.}
  \label{tab:runtime}
\end{table}

\begin{figure}
  \centering
  \begin{subfigure}{0.45\linewidth}
    \includegraphics[width=0.99\linewidth]{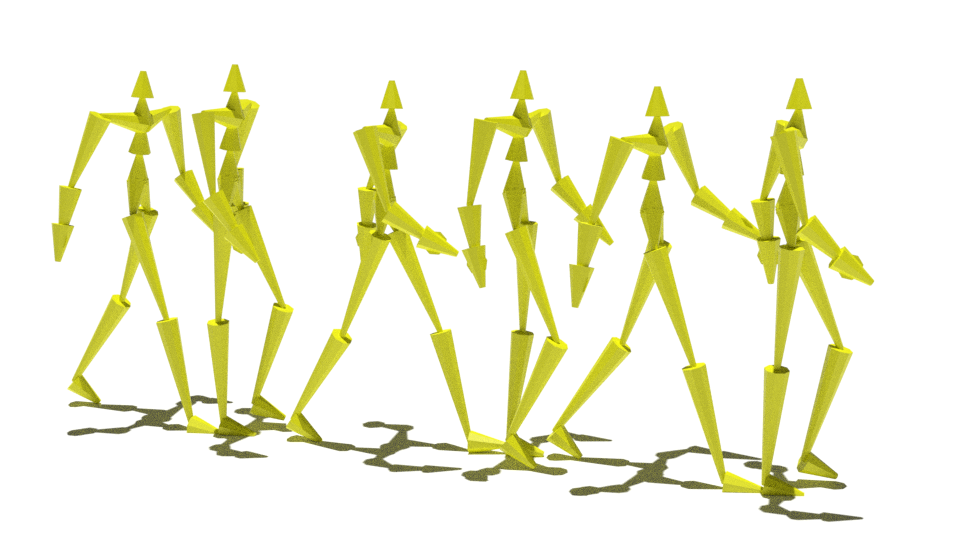}
    \caption{$\alpha$ = 0.0}
  \end{subfigure}
  \begin{subfigure}{0.45\linewidth}
    \includegraphics[width=0.99\linewidth]{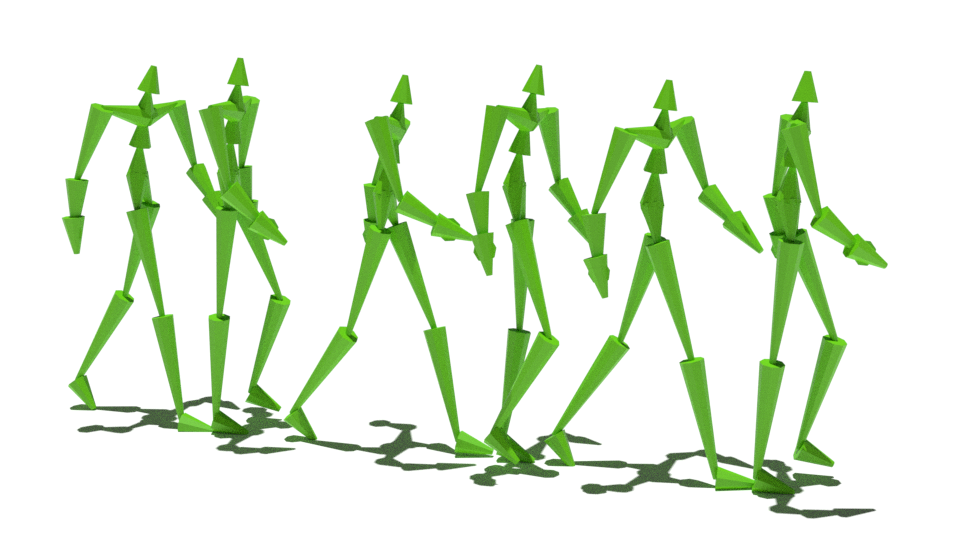}
    \caption{$\alpha$ = 0.3}
  \end{subfigure}
  \begin{subfigure}{0.45\linewidth}
    \includegraphics[width=0.99\linewidth]{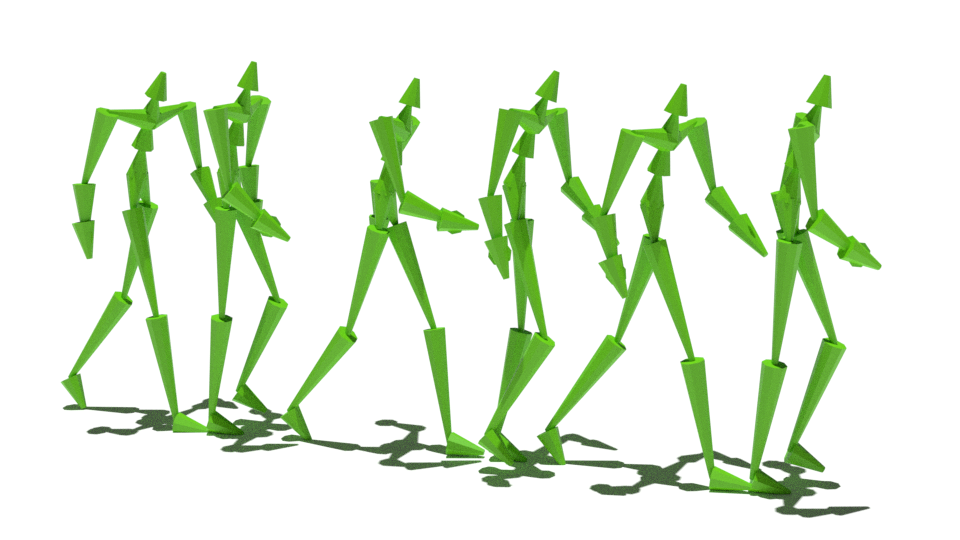}
    \caption{$\alpha$ = 0.6}
  \end{subfigure}
  \begin{subfigure}{0.45\linewidth}
    \includegraphics[width=0.99\linewidth]{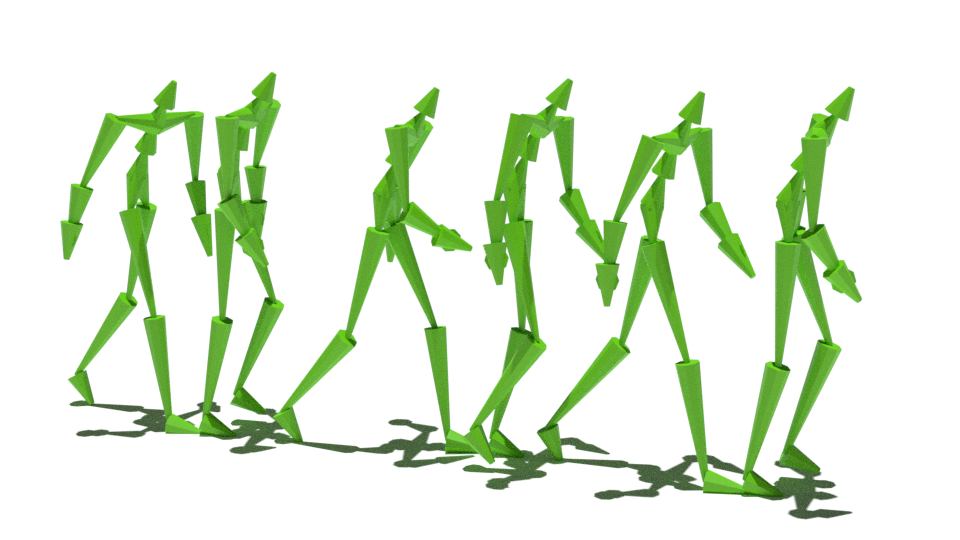}
    \caption{$\alpha$ = 1.0}
  \end{subfigure}
  \caption{Style interpolation. Task: Neutral walk into depressed style. $\alpha=0$ is equivalent to reconstruction of input motion.}
  \label{fig:interpolation}
\end{figure}
\subsection{Style Interpolation}
\label{ssec:style_interpolation}

\begin{figure*}
  \centering
  \footnotesize
  \begin{tabular}{cccc}
         Input Motion from \textit{Mixamo} & Ours & Aberman \etal~\cite{aberman2020unpaired} & Park \etal~\cite{park2021diverse}\\
         
         \midrule
         \midrule
         
        \begin{subfigure}{0.22\textwidth}
            \includegraphics[width=\linewidth]{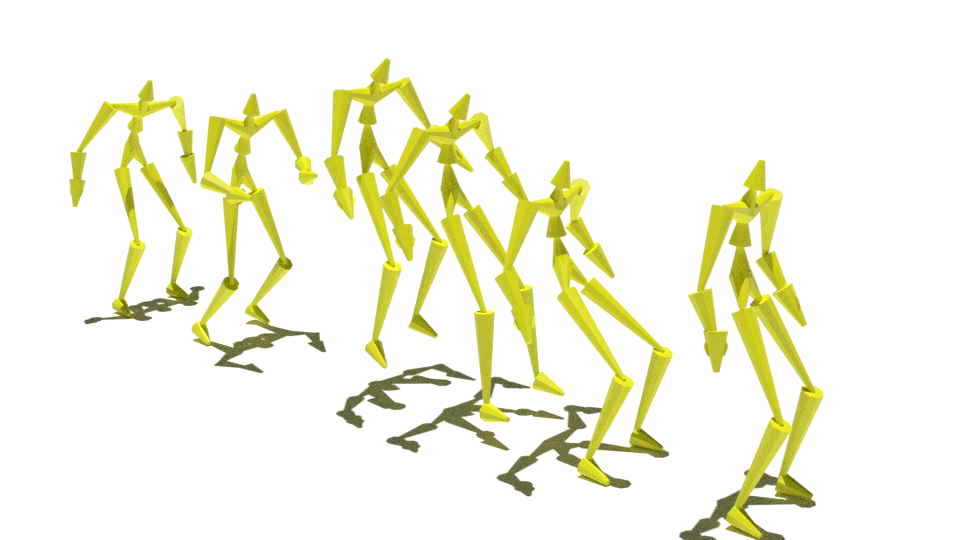}
        \end{subfigure}&
        \begin{subfigure}{0.22\textwidth}
            \includegraphics[width=\linewidth]{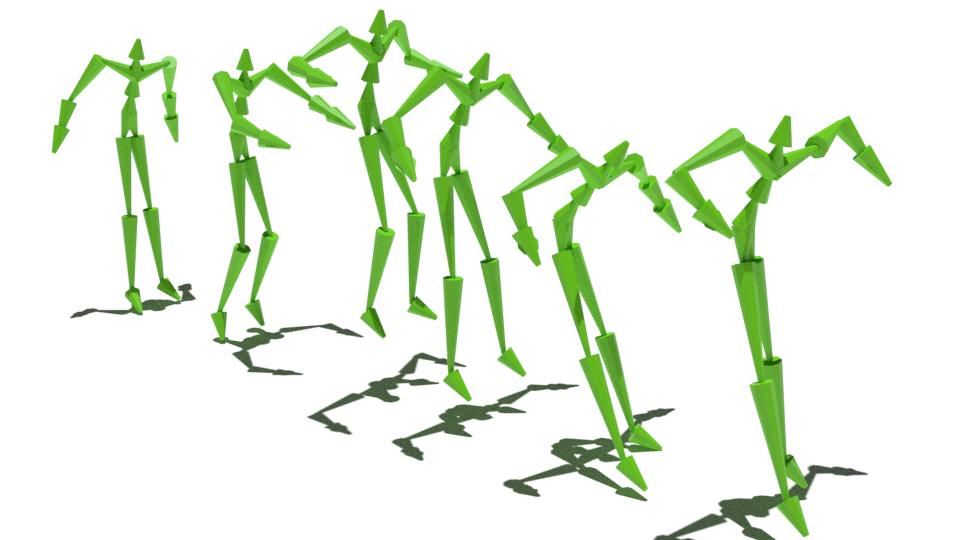}
        \end{subfigure}&
        \begin{subfigure}{0.22\textwidth}
            \includegraphics[width=\linewidth]{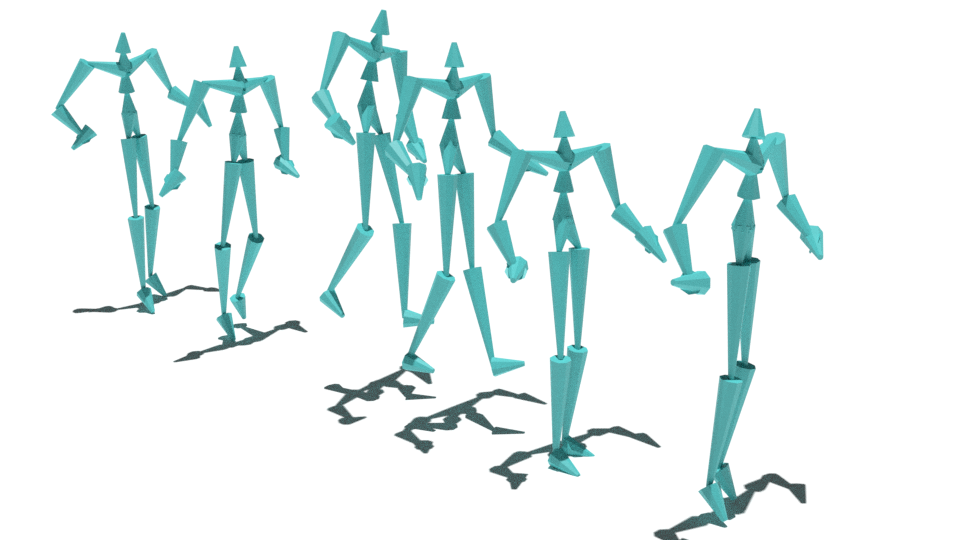}
        \end{subfigure}&
        \begin{subfigure}{0.22\textwidth}
            \includegraphics[width=\linewidth]{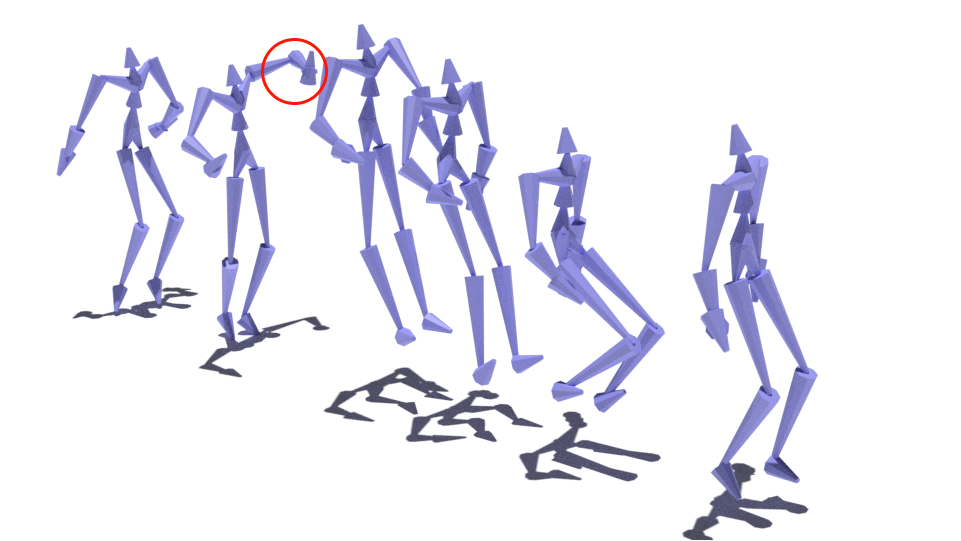}
        \end{subfigure}\\
        
        \multicolumn{4}{c}{(a)~Neutral jump into proud}\\
        
        \midrule
        
        \begin{subfigure}{0.22\linewidth}
            \includegraphics[width=\linewidth]{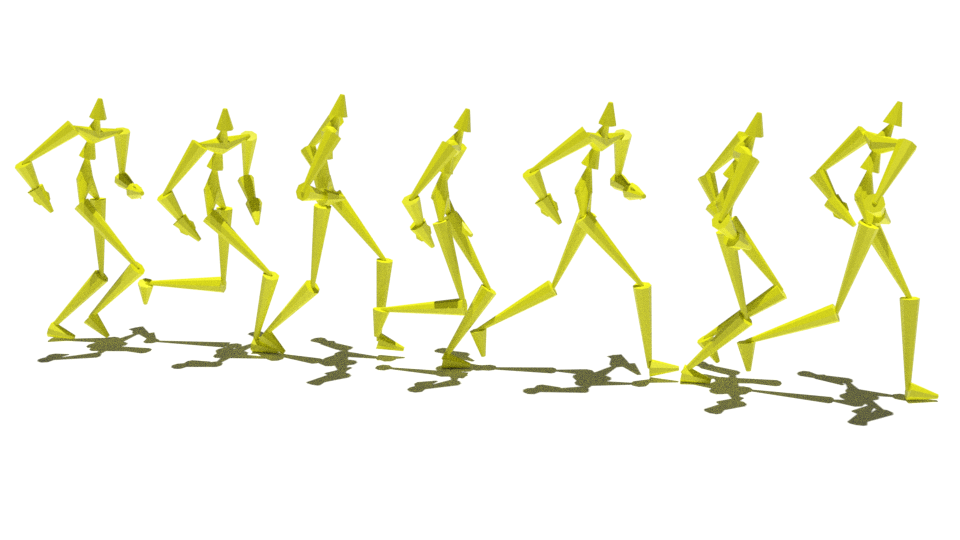}
        \end{subfigure}&
        \begin{subfigure}{0.22\linewidth}
            \includegraphics[width=\linewidth]{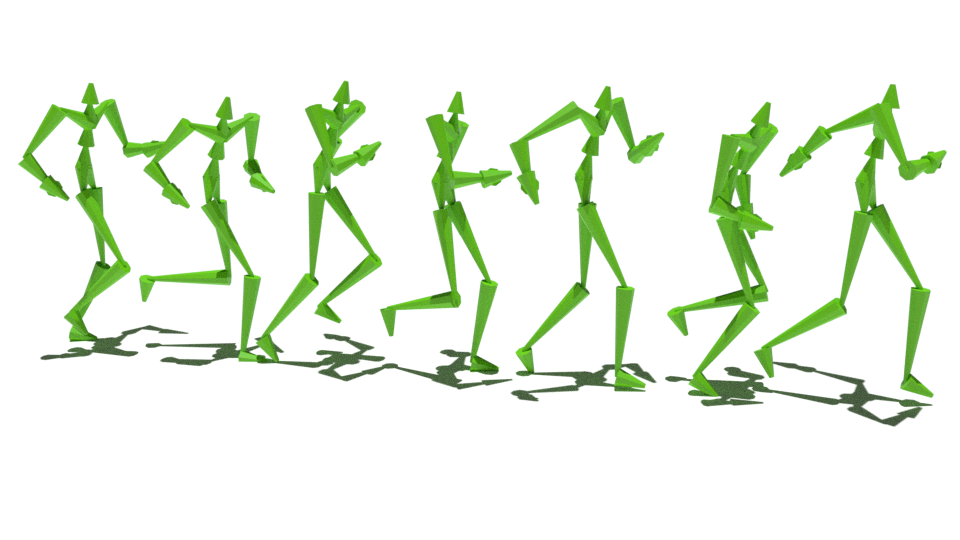}
        \end{subfigure}&
        \begin{subfigure}{0.22\linewidth}
            \includegraphics[width=\linewidth]{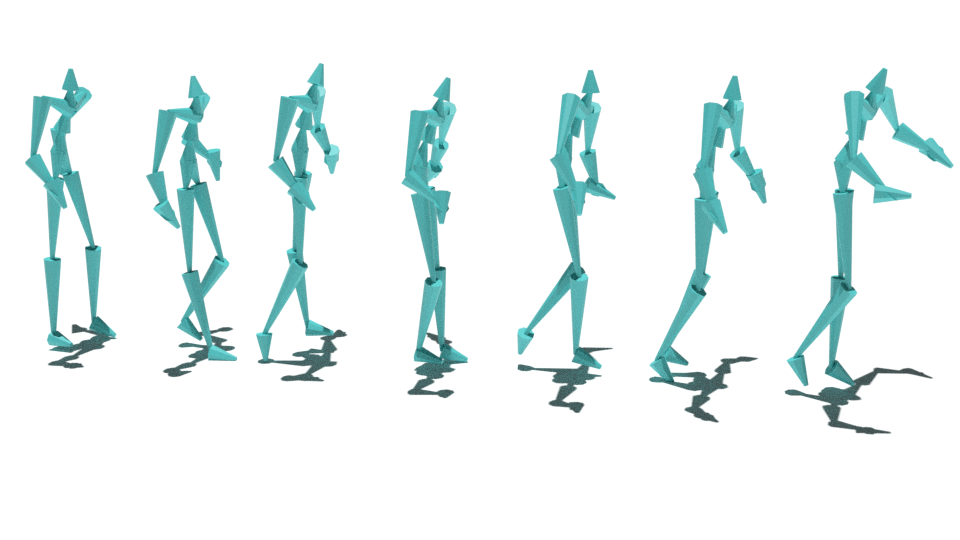}
        \end{subfigure}&
        \begin{subfigure}{0.22\textwidth}
            \includegraphics[width=\linewidth]{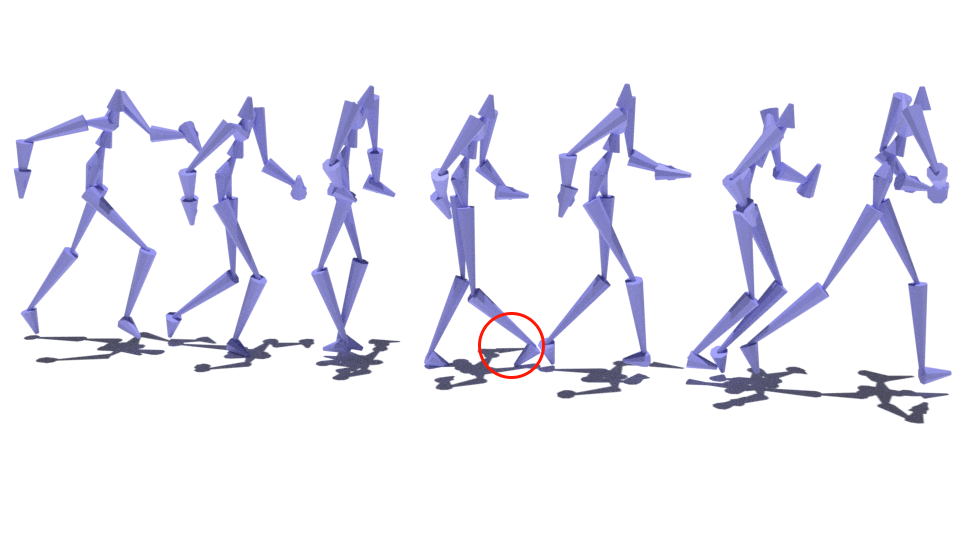}
        \end{subfigure}\\
        
        \multicolumn{4}{c}{(b)~Neutral run into angry}\\
        
  \end{tabular}
 \caption{Generalization test: style transfer results for our method and alternative methods~\cite{aberman2020unpaired,park2021diverse} on motion data from \textit{Mixamo}. \textit{Mixamo} motion data is unseen at training time and has different motion patterns. Some obvious artifacts are circled in red.}
 \label{fig:generalizability}
\end{figure*}
We also experiment with style interpolation. More specifically, we set the latent code $z_{t}'$ to be either a linear combination of two target styles $\hat{S}, \Tilde{S}$: 
$
    z_{t}' = r_{0}(z_{t}) + \alpha * r_{\hat{S}}(z_{t}) + (1 - \alpha) * r_{\Tilde{S}}(z_{t}),
    \label{eq:style_interpolation}
$
or a scaled value of a designated target style:
$
    z_{t}' = r_{0}(z_{t}) + \alpha * r_{\hat{S}}(z_{t}),
    \label{eq:style_mixture}
$
where $\alpha$ is a scalar in $[0, 1]$. As shown in ~Fig.~\ref{fig:interpolation}, motions with different style intensity can be produced by adjusting the coefficient $\alpha$ of a single target style. In the supplementary material, we demonstrate generating motions in a mixture of styles.

\subsection{Generalizability}
\label{ssec: robustness}

Motion capture data naturally contains significant variability due to motion speeds, the actor skeleton, \etc. The style transfer algorithm is likely to encounter out-of-distribution motion data. To examine robustness to motion perturbations, we test our framework and the approaches proposed in~\cite{aberman2020unpaired, park2021diverse} on retargeted public motion data from \emph{Mixamo}~\cite{mixamo}. As shown in Fig.~\ref{fig:generalizability}, our method can successfully transfer the unseen motion data to the desired style, while the two alternative methods contain apparent artifacts.


\section{Ablation Studies}
\label{sec:ablation}

We conduct extensive ablation studies to verify the relevance of multiple components in our framework, including each term in the generator loss $\mathcal{L}_{gen}$, the attention mechanism in \emph{FT-Att Discriminator}, and learned initial hidden states. Quantitative measurements based on FMD are listed in Tab.~\ref{tab:fmd}, which reveal the necessity of the modules in our framework. Qualitative visualization for each ablation experiment is included in the supplementary material.

\noindent\textbf{Which terms matter in the loss function?} The residual model has previously been proposed with a reconstruction loss to generate stylized motion by Mason \etal~\cite{mason2018few}. Therefore, we evaluate the need for the supervision modules, and set the training objective $\mathcal{L}_{gen}$ to be a pure regression task using $\mathcal{L}_{rec}$ in Eq.~\ref{eq: rec}. As a result, we find that \emph{Style-ERD} fails to produce motions in the target style when only trained on the reconstruction objective. Instead, the transferred motion is minimally different from the input motion.

Since the attention module in \emph{FT-Att Discriminator} depends on both content and target style labels, the discriminator can theoretically supervise both style and content of the motion. Thus, we remove the perceptual loss from the generator loss, \ie,~$\mathcal{L}_{gen} = \mathcal{L}_{rec} + \mathcal{L}_{adv}$. We find that some style transfer results experience obvious content changes after transfer without the content supervision module.

We also attempt to only remove the adversarial loss $\mathcal{L}_{adv}$ from the generator loss, \ie,~$\mathcal{L}_{gen} = \mathcal{L}_{rec} + \mathcal{L}_{per}$. Without the adversarial loss, the \emph{Style-ERD} model behaves similarly to an autoencoder, which can only reconstruct the input.

\noindent\textbf{Can we discard the attention mechanism in the discriminator?} We replace the proposed \emph{FT-Att Discriminator} with a standard multi-class discriminator used in~\cite{aberman2020unpaired,liu2019few}. The multi-class discriminator has a shared convolutional feature extractor and individual heads for each style discrimination task. We find that the full model produces results with more expressive styles while the baseline discriminator fails to capture some style features and contains artifacts.

\noindent\textbf{Do learnable initial hidden states help?} Instead of setting the hidden states as learnable parameters, the initial hidden states of LSTM layers are set to zero as done in~\cite{li2017auto}. We find that 
the transfer results can exhibit a temporal shift, especially for short motion clips. For example, the start of a motion can be transferred to the middle stage of the motion sequence.


\section{Conclusion}
\label{sec:con}

In summary, we introduce a novel style transfer model, \emph{Style-ERD}, with the Encoder-Recurrent-Decoder structure as a solution to the online motion style transfer problem. In our style-modeling framework, the memory module encapsulates the style and content context of the past frames. We incorporate learnable initial hidden states conditioned on the input to enhance the responsiveness of our method. Furthermore, we propose a new discriminator \emph{FT-Att Discriminator} with attention on both feature and temporal dimensions to supervise the style of the output. Our method enables stylizing the input frame with a minimal delay while significantly accelerating the transfer process in an online setting. Compared with previous methods, our \emph{Style-ERD} model is able to produce more realistic style transferred motions while being robust to perturbations in the input. \update{We discuss the limitations and societal impact of our work in the supplementary material.}



{\small
\bibliographystyle{ieee_fullname}
\bibliography{egbib}
}

\end{document}


\title{Style-ERD: Responsive and Coherent Online Motion Style Transfer Supplementary Material
}  

\maketitle
\thispagestyle{empty}

\appendix

\section{Implementation Details}

Our method is implemented in PyTorch and trained on a PC with an 8-core AMD Ryzen 7 CPU with a single NVIDIA GeForce GTX 1060 GPU. The overall training process takes about 10 hours to complete. Both the generator and discriminator are optimized by Adam optimizer \cite{kingma2014adam} with learning rates of $10^{-4}$ and $5 \times 10^{-5}$ respectively. Our model is trained for 2000 epochs with a batch size of 16. We adopt the default initialization offered by PyTorch for all the neural networks in our model.

The encoder in the style transfer module consists of two MLP layers with $(64, 32)$ neurons and ReLU activation. The latent code $z$ has a dimension of 32. The residual module comprises 6 LSTM layers for the neutral branch $r_{0}$ and 4 LSTM layers for other branches $[r_{1},\dots,r_{n_{S}}]$. The decoder is conditioned on target style labels and comprises 4 MLP layers with $(58, 86, 128, 184)$ neurons and ReLU activation. For the discriminator, we find that replacing ReLU with LeakyReLU accelerates the training.

We experiment with two different loss functions on the quaternion values: L2-norm and $\mathcal{L}_{quat}$ in Eq.~(2). Though the four coefficients of a quaternion are continuous and smooth, the rotation value is not evenly-spaced. Optimizing in this not evenly-spaced space with L2-norm can lead to jerky motion, which can be solved by adopting $\mathcal{L}_{quat}$.

We pre-train a denoising autoencoder with 1D convolution layers as the feature extractor for Frechet Motion Distance (FMD). At training time, the joint rotations are sampled from the dataset, and we add random noise sampled from a normal distribution, $\mathcal{N}(0, 0.03)$ to the joint rotations. The denoising autoencoder is trained to reconstruct the noisy joint rotation input with $\mathcal{L}_{quat}$ in Eq.~(2) as the loss function. After the training, we use the activation of the last convolution layer in the encoder as features for the FMD score.

Code, models and demo videos are available at \href{https://tianxintao.github.io/Online-Motion-Style-Transfer}{https://tianxintao.github.io/Online-Motion-Style-Transfer}.

\section{User Study}

We performed a user study to compare the quality of the style transfer results. We presented the participants with the rendered motions in a side view angle. 

The questionnaire contained eight sets of transfer results in the target style and also the input motion for reference. Each set contained the motions generated by \emph{Style-ERD} and alternative methods~\cite{aberman2020unpaired,park2021diverse}. The users were supposed to select the best results in three aspects: style expressiveness, content preservation and overall quality.

\begin{table}
  \centering
  \begin{tabular}{l|c|c|c}
    \toprule
    Method & Style & Content & Quality\\
    \midrule
    Aberman \etal~\cite{aberman2020unpaired} & 9.5\% & 4.5\% & 6.1\%\\
    Park \etal~\cite{park2021diverse} & 12.3\% & 8.6\% & 5.8\%\\
    Ours & $\textbf{78.2\%}$ & $\textbf{86.8\%}$ & $\textbf{88.1\%}$\\
    \bottomrule
  \end{tabular}
  \caption{User study results. Each entry in the table corresponds to the ratio of being selected as the optimal among the alternatives in the three aspects.
  }
  \label{tab:user}
\end{table}

We collected a total of 243 responses from 31 participants. As shown in Tab.~\ref{tab:user}, $78.2\%$, $86.8\%$ and $88.1\%$ of the responses select our results as the best among all the choices in the three aspects respectively. The results of the user study suggest that \emph{Style-ERD} outperforms the alternative methods in terms of the style transfer quality.
\section{Discussions}


\subsection{Limitations}
Our method still requires paired motion data in different styles. This could possibly be addressed by adopting the cycle consistency idea~\cite{zhu2017unpaired}. Also, our model is conditioned on content labels, which could be unavailable. 
Additionally, \emph{Style-ERD} has an increasing runtime with a growing input length.
As with other approaches, we still rely on inverse kinematics based cleanup to remove minor foot sliding.
As future work, we are also interested in optimizing our framework to quickly adapt to unseen styles in a few-shot by matrix decomposition as shown in~\cite{mason2018few}.

\subsection{Societal Impact}
The motion data is biased towards one default skeleton, ignoring the needs of minority groups, \eg,~children, elders and people with physical disabilities. This can be mitigated by motion retargetting in the pre-processing step. Motion style transfer could be used for creating human animations with the intention to mislead.

\subsection{Style Transfer with Unpaired Data}
\label{ssec: heterogeneous}
We design our framework to perform motion style transfer with paired motion data. For example, to transfer a neutral walking motion to angry, the dataset should contain an angry walking motion as reference, which does not need to be temporally registered. However, some researchers have demonstrated style transfer with unpaired motion data is also feasible~\cite{aberman2020unpaired}. Such style transfer with unpaired data is defined as heterogeneous style transfer. Exploring style transfer with unpaired data remain as a future work since it lowers the requirement on the dataset and enables to learn more diverse style transfer effects. 

\subsection{Choice of Input Style}
\begin{figure}
  \centering
  \begin{subfigure}{0.49\linewidth}
    \includegraphics[width=0.95\linewidth]{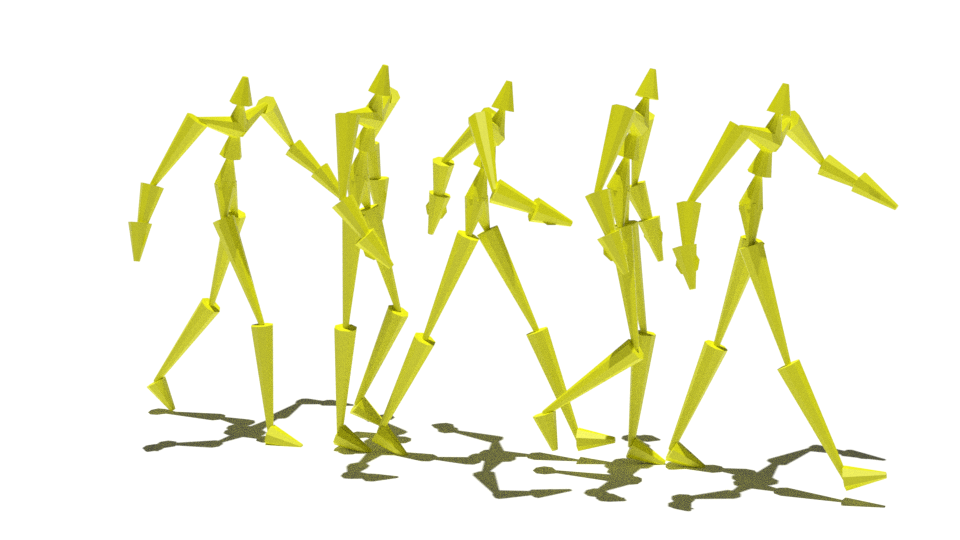}
    \caption{Input: Angry walk}
  \end{subfigure}
  \begin{subfigure}{0.49\linewidth}
    \includegraphics[width=0.95\linewidth]{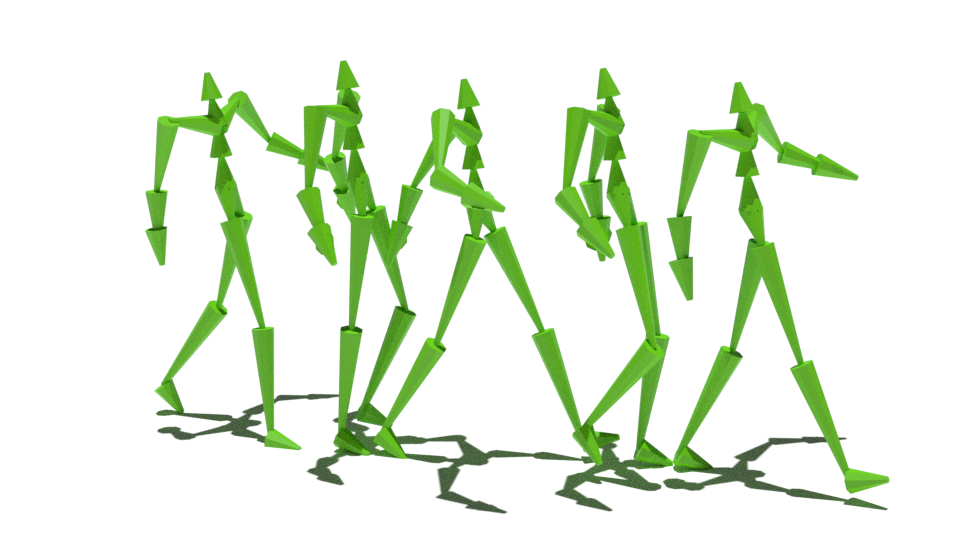}
    \caption{Output: Strutting walk}
  \end{subfigure}
  \caption{Style transfer with non-neutral input motion. Our model is not restricted by the input motion style.}
  \label{fig:inputstyle}
\end{figure}

\begin{figure}
  \centering
  \begin{subfigure}{0.49\linewidth}
    \includegraphics[width=0.95\linewidth]{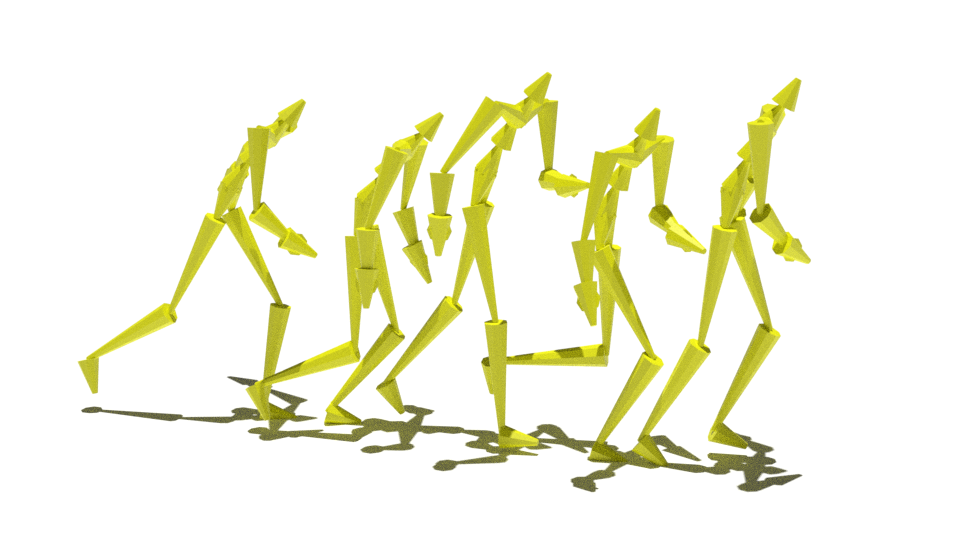}
    \caption{Input: Depressed run}
  \end{subfigure}
  \begin{subfigure}{0.49\linewidth}
    \includegraphics[width=0.95\linewidth]{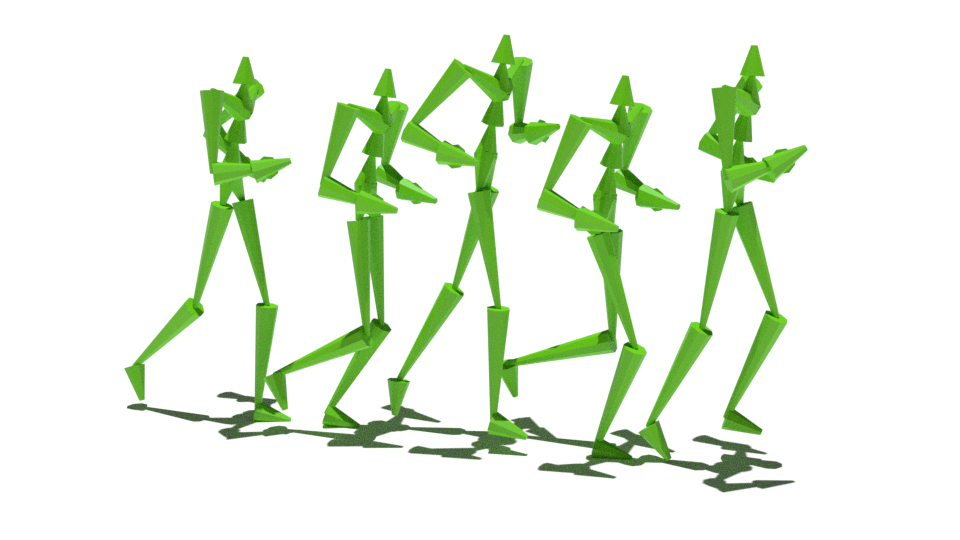}
    \caption{Output: Neutral run}
  \end{subfigure}
  \caption{Transfer non-neutral input to neutral style.}
  \label{fig:toneutral}
\end{figure}

\begin{figure}
  \centering
  \begin{subfigure}{0.99\linewidth}
    \centering
    \includegraphics[width=0.99\linewidth]{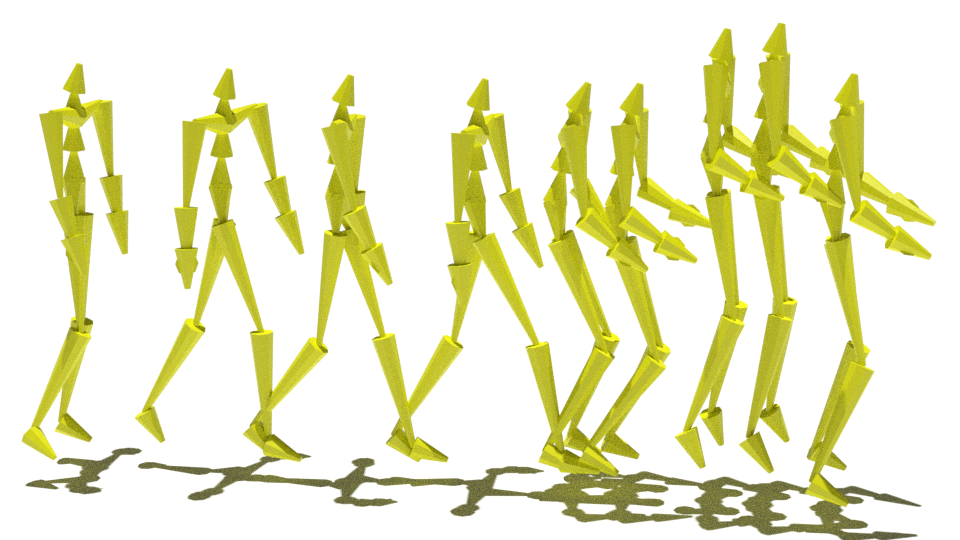}
  \end{subfigure}
  \begin{subfigure}{0.99\linewidth}
    \centering
    \includegraphics[width=0.99\linewidth]{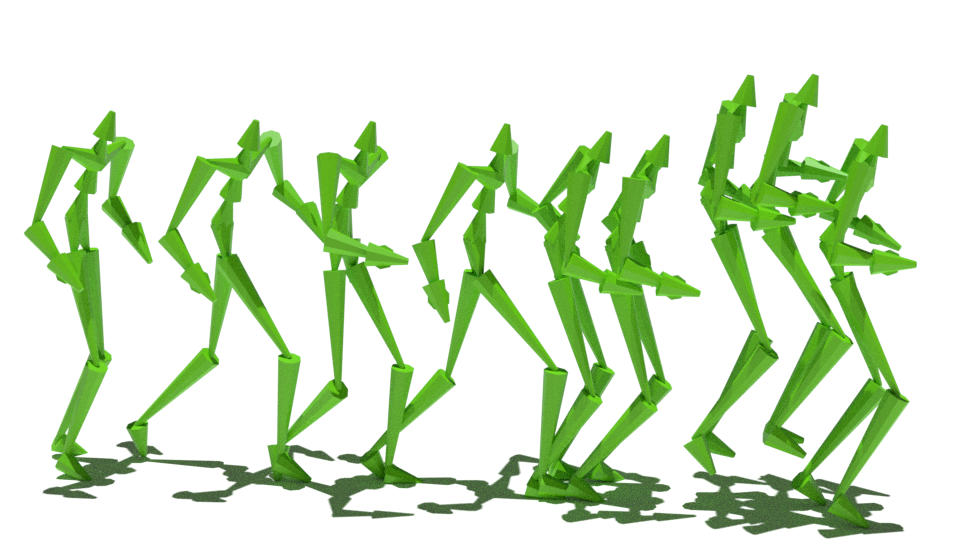}
  \end{subfigure}
  \caption{Style transfer on heterogeneous motion sequence. Task: Neutral walk and jump motion to old.}
  \label{fig:hetergeneous}
\end{figure}

\begin{figure}
  \centering
  \begin{subfigure}{0.98\linewidth}
    \includegraphics[width=0.95\linewidth]{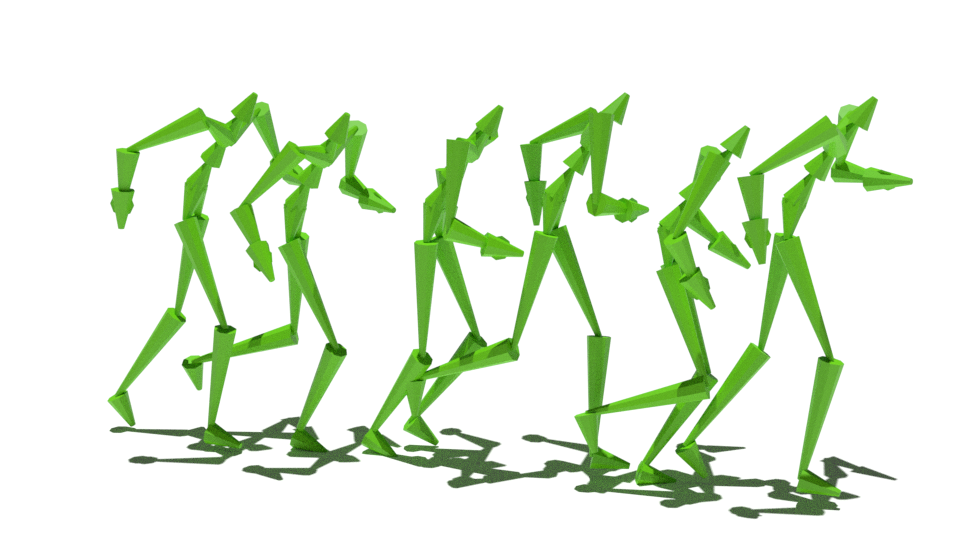}
    \caption{Neutral run to old and depressed ($\alpha$ = 0.5)}
  \end{subfigure}
  \begin{subfigure}{0.98\linewidth}
    \includegraphics[width=0.95\linewidth]{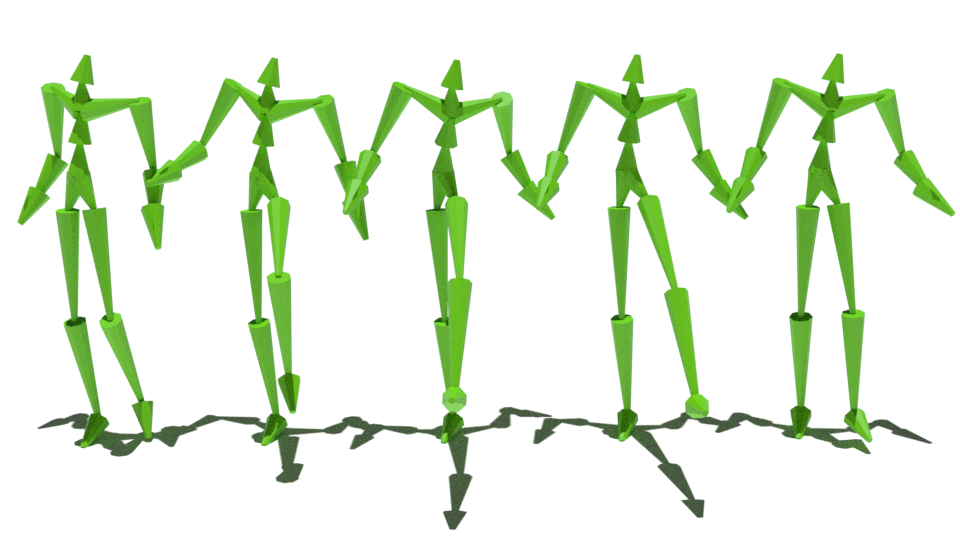}
    \caption{Neutral kick to childlike and strutting ($\alpha$ = 0.5)}
  \end{subfigure}
  \caption{Mixture of two existing styles.}
  \label{fig:mixture}
\end{figure}


\begin{figure}[t]
  \centering
  \begin{subfigure}{\linewidth}
      \begin{subfigure}{0.49\linewidth}
          \includegraphics[width=\linewidth]{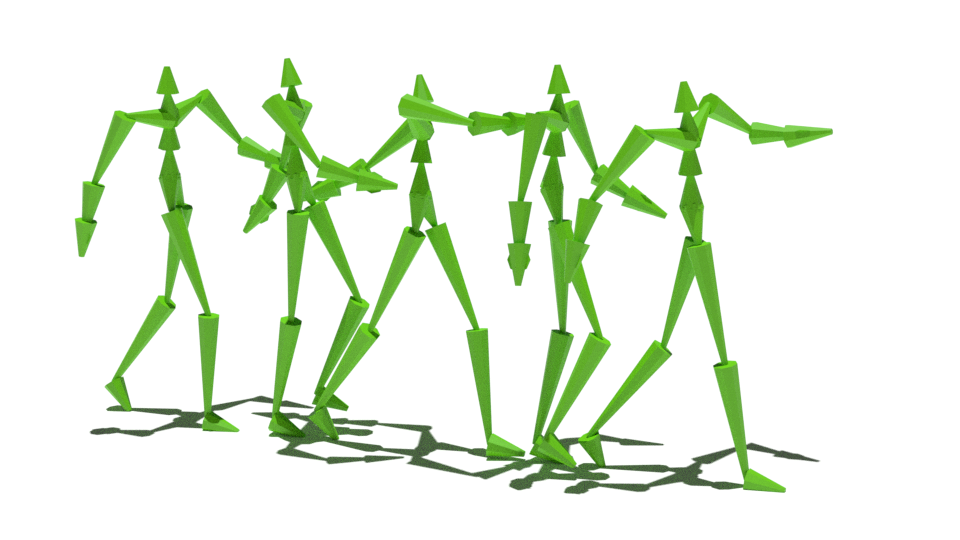}
      \end{subfigure}
      \begin{subfigure}{0.49\linewidth}
          \includegraphics[width=\linewidth]{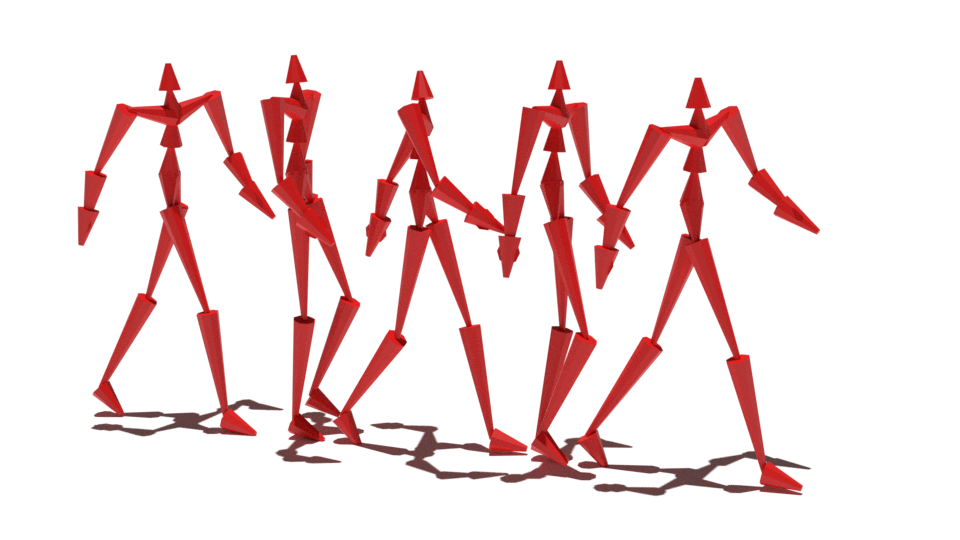}
      \end{subfigure}
      \caption{$\mathcal{L}_{gen}=\mathcal{L}_{rec}$. Task: Neutral walk to proud.}
      \label{fig:abalation_reconstruction}
  \end{subfigure}
  \begin{subfigure}{\linewidth}
      \begin{subfigure}{0.48\linewidth}
        \includegraphics[width=\linewidth]{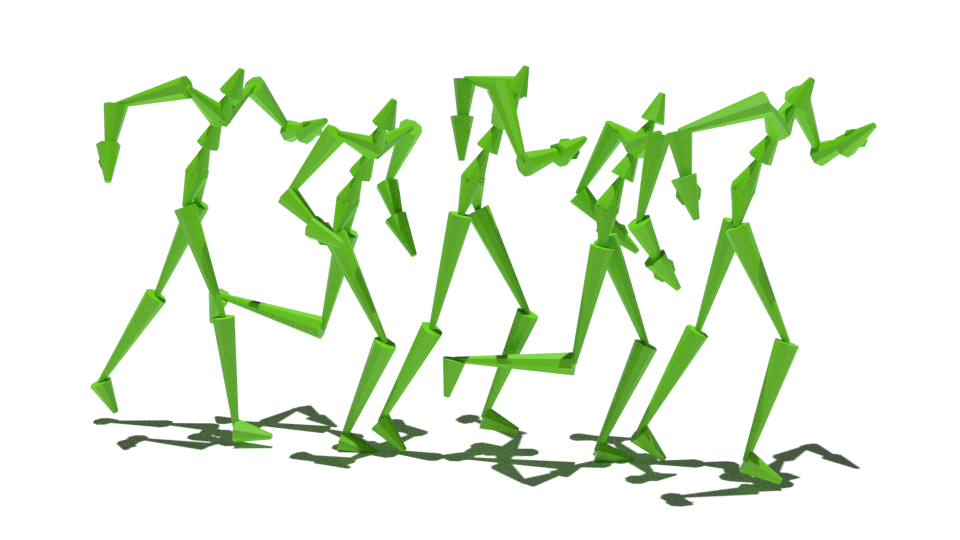}
      \end{subfigure}
      \begin{subfigure}{0.48\linewidth}
        \includegraphics[width=\linewidth]{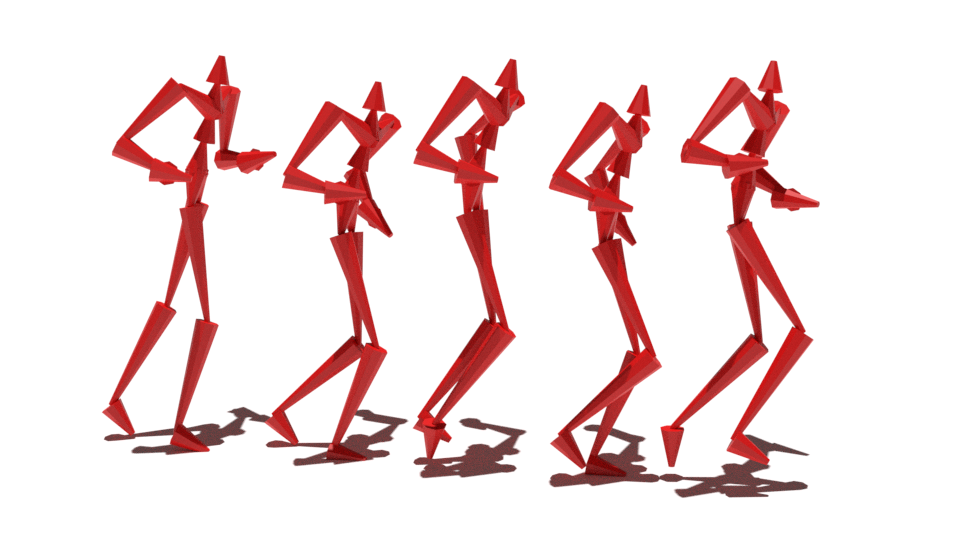}
      \end{subfigure}
      \caption{$\mathcal{L}_{gen}=\mathcal{L}_{rec}+\mathcal{L}_{adv}$. Task: Neutral run to angry.}
      \label{fig:abalation_perceptual}
  \end{subfigure}
  \begin{subfigure}{\linewidth}
      \begin{subfigure}{0.49\linewidth}
          \includegraphics[width=\linewidth]{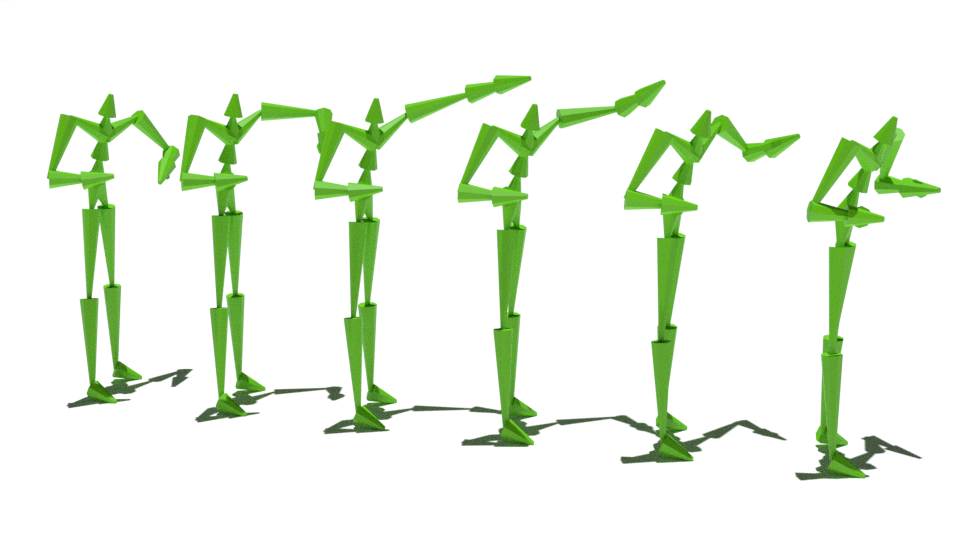}
      \end{subfigure}
      \begin{subfigure}{0.49\linewidth}
          \includegraphics[width=\linewidth]{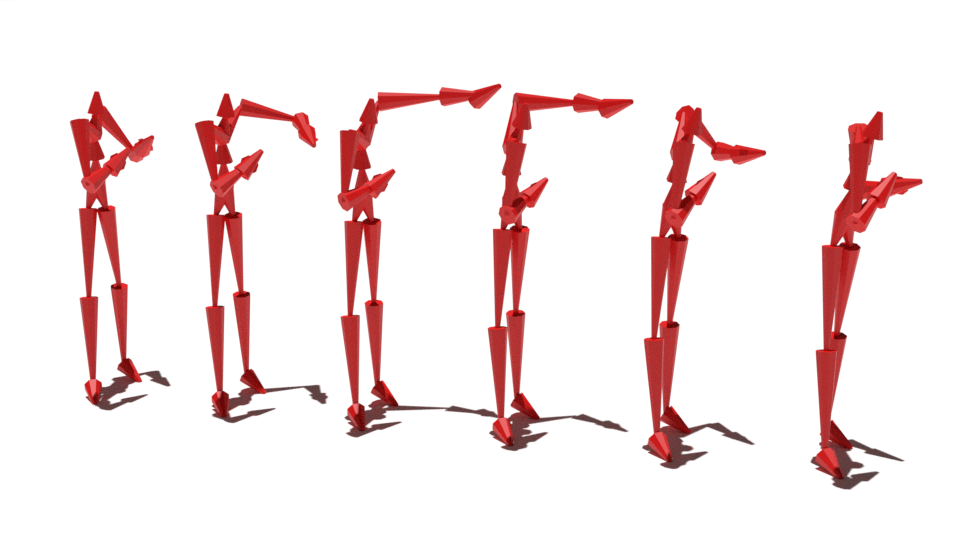}
      \end{subfigure}
      \caption{$\mathcal{L}_{gen}=\mathcal{L}_{rec}+\mathcal{L}_{per}$. Task: Neutral punch to depressed.}
      \label{fig:abalation_adv}
  \end{subfigure}
  \begin{subfigure}{\linewidth}
      \begin{subfigure}{0.49\linewidth}
        \includegraphics[width=\linewidth]{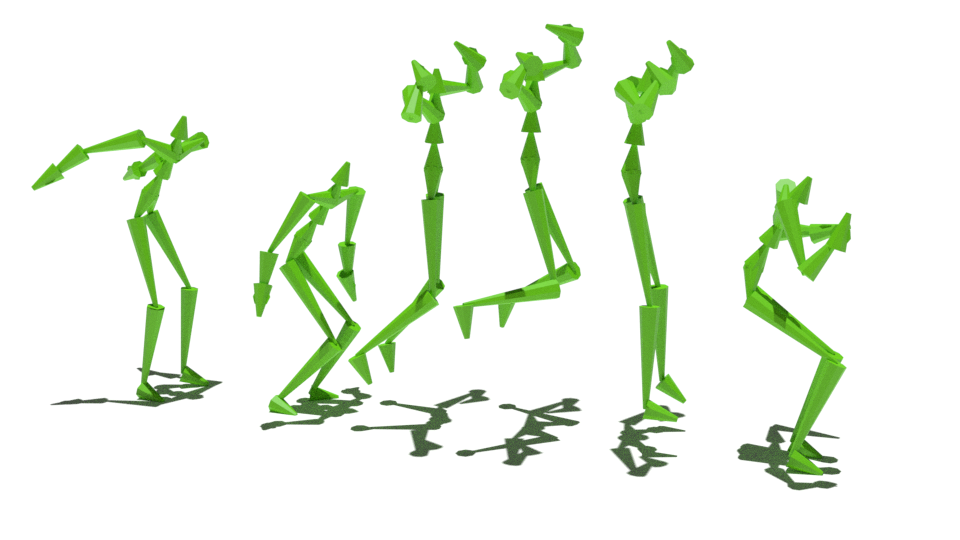}
      \end{subfigure}
      \begin{subfigure}{0.49\linewidth}
        \includegraphics[width=\linewidth]{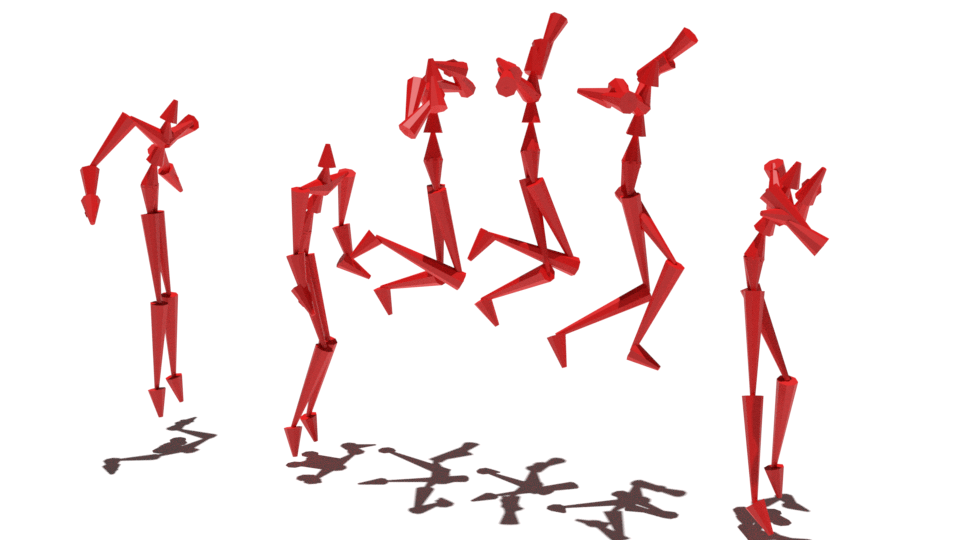}
      \end{subfigure}
      \caption{Attention mechanism of the discriminator. Task: Neutral jump to sexy.}
      \label{fig:ablation_attention}
  \end{subfigure}
  \begin{subfigure}{\linewidth}
      \begin{subfigure}{0.49\linewidth}
        \includegraphics[width=\linewidth]{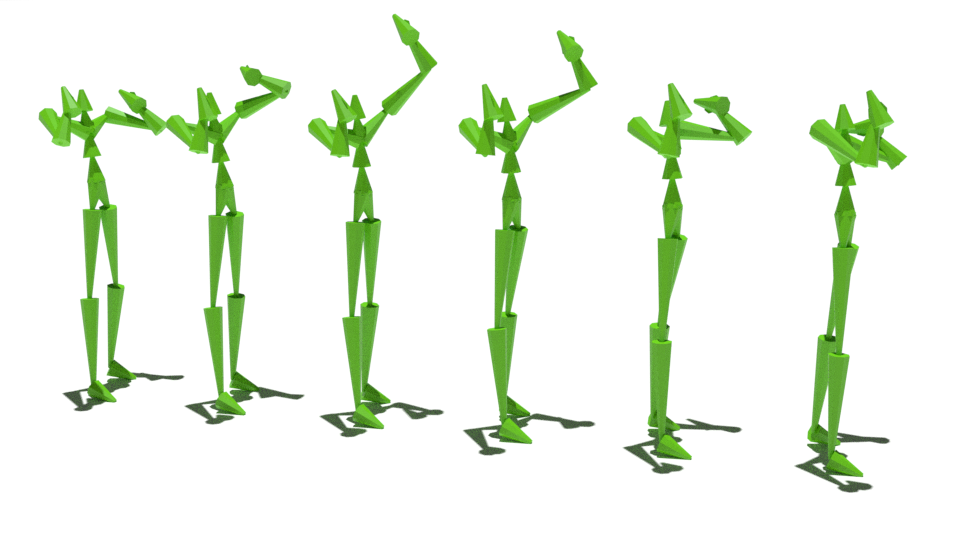}
      \end{subfigure}
      \begin{subfigure}{0.49\linewidth}
        \includegraphics[width=\linewidth]{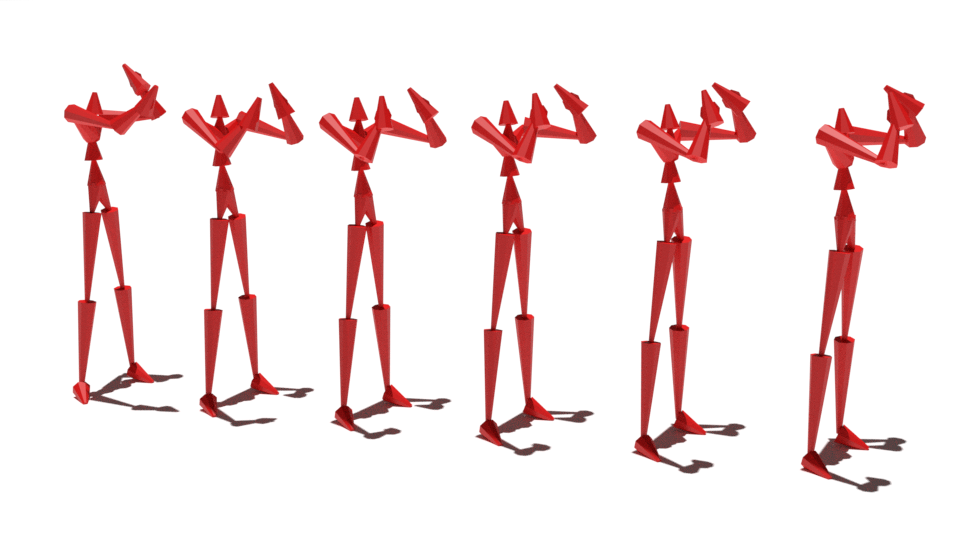}
      \end{subfigure}
      \caption{Learnable initial hidden states. Task: Neutral punch to childlike.}
      \label{fig:abalation_initial_sates}
  \end{subfigure}
  \caption{Ablation studies. Left (green): results produced by the full model, Right (red): ablation experiment results.}
  \label{fig:ablation}
\end{figure}

Most qualitative results in this work use input motion in neutral style. Style transfer starting from neutral motion is a more practical problem because more neutral motion exists in the available database. Nevertheless, our framework is not restricted by the input style to be neutral. Fig.~\ref{fig:inputstyle} shows our method can successfully transfer input motion in angry style to motion in strutting style. Given the training process, transferring non-neutral motion to another style other than neutral is an unseen task at inference time. Moreover, we also experiment with transferring stylized motion to its neutral counterpart, and show the results in Fig.~\ref{fig:toneutral}. Such results reveal that the encoder $E$ in the style transfer module \emph{Style-ERD} can normalize input motion to neutral style as expected.

When the input motion is not clearly depicted with a style label, the motion can usually be considered neutral for the transfer algorithm and produces satisfying style transfer results. For example, the \emph{Mixamo} data we used in the generalization test does not own a specific style label. We treat it as neutral motion and feed it to the \emph{Style-ERD} module to generate the motions shown in Fig.~7 of the paper. Alternatively, a style-classification model can be trained to predict the style of the input motion. Therefore, \emph{Style-ERD} can accept the style prediction of the classification model as the input style label for the transfer task.
     


\section{Extra Visualization Results}

In this section, we show the results that are omitted in the paper, including style transfer results on heterogeneous motion sequence in Fig.~\ref{fig:hetergeneous}, mixture of two styles in Fig.~\ref{fig:mixture}, and ablation study results in Fig.~\ref{fig:ablation}.



{\small
\bibliographystyle{ieee_fullname}
\bibliography{egbib}
}